\documentclass{article}

\usepackage[main, final]{neurips_2026}
\workshoptitle{RLxF: Reinforcement Learning from World Feedback}

\usepackage[utf8]{inputenc} 
\usepackage[T1]{fontenc}    
\usepackage{xcolor}         
\definecolor{tggreen}{HTML}{4A7C59}
\usepackage[
    colorlinks=true,
    linkcolor=blue!50!black,
    citecolor=blue!50!black,
    urlcolor=blue!50!black
]{hyperref}
\usepackage{url}            
\usepackage{booktabs}       
\usepackage{amsfonts}       
\usepackage{nicefrac}       
\usepackage{microtype}      
\usepackage{subcaption}
\usepackage{fancyvrb}

\usepackage{multirow}
\usepackage{enumitem}
\usepackage{wrapfig}
\usepackage{bm}
\usepackage{algorithm}
\usepackage{algorithmic}
\usepackage{caption}
\usepackage{listings} 

\usepackage{amsmath}
\usepackage{amssymb}
\usepackage{mathtools}
\usepackage{amsthm}
\usepackage{tikz}
\usetikzlibrary{matrix}

\usepackage[capitalize,noabbrev]{cleveref}

\usepackage{tcolorbox}
\usepackage{float} 

\usepackage{tikz}
\usetikzlibrary{matrix,backgrounds}
\definecolor{cbaseline}{HTML}{1D6996}
\definecolor{cfresh}{HTML}{0F8554}
\definecolor{cper}{HTML}{E17C05}
\definecolor{c100k}{HTML}{EDAD08}
\definecolor{cere}{HTML}{FBC5F9}
\definecolor{cneutral}{HTML}{666666}
\newcommand{\swatch}[1]{\textcolor{#1}{\rule{5pt}{5pt}}\hspace{3pt}}

\tcbset{
  takeawaybox1/.style={
    colback=gray!10!white,
    colframe=gray!60!black,
    fonttitle=\bfseries\small,
    title=Takeaway 1 -- Replay Volume Matters,
    boxrule=0.8pt,
    arc=3pt,
    left=4pt, right=4pt, top=3pt, bottom=3pt,
  }
}
\tcbset{
  takeawaybox2/.style={
    colback=gray!10!white,
    colframe=gray!60!black,
    fonttitle=\bfseries\small,
    title=Takeaway 2 -- Sampling Entropy Matters,
    boxrule=0.8pt,
    arc=3pt,
    left=4pt, right=4pt, top=3pt, bottom=3pt,
  }
}

\tcbset{
  takeawaybox3/.style={
    colback=gray!10!white,
    colframe=gray!60!black,
    fonttitle=\bfseries\small,
    title=Summary of the Experimental Results,
    boxrule=0.8pt,
    arc=3pt,
    left=4pt, right=4pt, top=3pt, bottom=3pt,
  }
}

\newcommand{\volumename}{replay volume}
\newcommand{\Volumename}{Replay Volume}
\newcommand{\densityname}{expected recency}
\newcommand{\Densityname}{Expected Recency}
\newcommand{\entropyname} {sampling entropy}

\theoremstyle{plain}

\theoremstyle{definition}

\theoremstyle{remark}

\def\<#1,#2>{\langle #1,#2\rangle}

\NewDocumentCommand{\Var}{somo}{\mathrm{Var}\IfValueT{#2}{_{#2}}{} \IfBooleanTF{#1}{#3}{\IfValueTF{#4}{\!\left(#3\ \middle|\ #4\right)}{\parentheses*{#3}}}}

\title{When Does Non-Uniform Replay Matter \\ in Reinforcement Learning?}

\author{%
  Michal Korniak$^{\dagger\;1,2}$\thanks{Corresponding author: \texttt{michael.korniak@gmail.com} \\ \phantom{Corr}Code available at \url{https://github.com/M-Korniak/when-does-replay-matter} \\
  \phantom{Corr}Project page at \url{https://m-korniak.github.io/when-does-replay-matter}} \quad
  Mikołaj Czarnecki$^{\dagger\;2}$ \quad
  Yarden As$^{1}$ \\
  \textbf{Piotr Miłoś$^{2}$} \quad
  \textbf{Pieter Abbeel$^{3,4}$} \quad
  \textbf{Michal Nauman$^{2,3}$} \\[0.5em]
  $^1$ETH Zurich \quad
  $^2$University of Warsaw \quad
  $^3$UC Berkeley \quad
  $^4$Amazon FAR \\[0.3em]
  $^\dagger$Equal contribution
}

\begin{document}

\maketitle

\begin{abstract}
\looseness=-1Modern off-policy reinforcement learning algorithms often rely on simple uniform replay sampling and it remains unclear \emph{when and why} non-uniform replay improves over this strong baseline. Across diverse RL settings, we show that the effectiveness of non-uniform replay is governed by three factors: \emph{replay volume}, the number of replayed transitions per environment step; \emph{expected recency}, how recent sampled transitions are; and the \emph{entropy} of the replay sampling distribution. Our main contribution is  clarifying when non-uniform replay is beneficial and providing practical guidance for replay design in modern off-policy RL. Namely, we find that non-uniform replay is most beneficial when replay volume is low, and that high-entropy sampling is important even at comparable expected recency. Motivated by these findings, we adopt a simple Truncated Geometric replay that biases sampling toward recent experience while preserving high entropy and incurring negligible computational overhead. Across large-scale parallel simulation, single-task, and multi-task settings, including four modern and classic algorithms evaluated on six RL benchmark suites, this replay sampling strategy improves sample efficiency in low-volume regimes while remaining competitive when replay volume is high. 
\end{abstract}
\section{Introduction}
\label{sec:introduction}

Experience replay is one of the central components of off-policy RL~\citep{mnih2015human, silver2017mastering}. Many replay sampling strategies have been proposed, from uniform replay to distributions biased by error~\citep{Schaul2015} or recency~\citep{wang2019boostingsoftactorcriticemphasizing}. Yet recent continuous-control algorithms still often use uniform replay by default~\citep{bhatt2024crossq,nauman2024bigger,SimbaV2,palenicek2025xqc}. Understanding when this choice is sufficient, and when it can be improved, is especially important in regimes such as parallel simulation, multi-task learning, and other settings where data collection outpaces optimization. In these regimes, individual transitions may be replayed only a few times, so the replay sampling distribution can have a larger effect on learning.

A natural way to analyze replay is to count how often each transition is reused over the course of training. This whole-training exposure view underlies prior analyses of recency-biased replay~\citep{wang2019boostingsoftactorcriticemphasizing,wang2020striving}: changing the sampler changes how replay updates are distributed across transitions over their lifetime. However, whole-training exposure conflates two distinct factors. A learner can consume more recent data either by shifting the sampling distribution toward newer transitions, or by increasing the number of replayed transitions per environment step through a larger update-to-data ratio (UTD) or batch size. Moreover, whole-training exposure does not fully explain empirical behavior observed in off-policy RL. In runs where replay capacity is comparable to the training horizon, exposure can be highly skewed, yet non-uniform replay often provides little benefit~\citep{nauman2024bigger, SimbaV2}. Conversely, as we show in this paper, in high-throughput simulation, exposure can be nearly flat throughout training, yet recency-biased replay can substantially outperform uniform sampling. These observations suggest that whole-training exposure alone does not predict when non-uniform replay improves performance.

\begin{figure}[t!]
    \centering
    \vspace{-0.05in}
    \includegraphics[width=\textwidth]{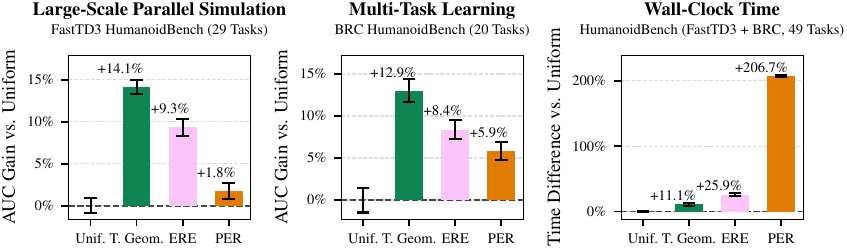}
    \vspace{-0.2in}
    \caption{\emph{\textbf{ Performance and runtime trade-offs on HumanoidBench.}}
    Relative sample efficiency gains measured as area under (learning) curve (AUC) for large-scale parallel simulation FastTD3 (\textit{left}) and multi-task BRC (\textit{middle}), and wall-clock time difference aggregated across both settings (\textit{right}) compared to uniform sampling. Error bars show $95\%$ stratified bootstrap CI. We report results in large-scale parallel FastTD3 and multi-task BRC because they feature relatively low experience reuse per transition, where non-uniform replay is expected to have the greatest impact (Subsection~\ref{sec:sampling_scale}). The recency-biased truncated geometric (described in detail in~\Cref{sec:truncated-geometric}) sampling yields consistent gains with minimal overhead, while PER exhibits higher computational load. Experiment details in~\Cref{sec:experiments}. We refer to~\Cref{app:auc} for discussion on AUC as a suitable metric for \emph{sample efficiency}.}
    \label{fig:frontpage}
    \vspace{-0.75cm}
\end{figure}

In this paper, we expand on this perspective by showing that replay performance also depends on factors that whole-training exposure does not account for. This view separates three factors: \textbf{\densityname} captures how recent sampled transitions are on average; \textbf{\volumename}, given by UTD times batch size, captures how many replayed transitions are consumed between consecutive environment steps; and  \textbf{sampling entropy} captures how concentrated the replay sampling distribution is at each update. This decomposition separates changing the sampler from changing the amount of optimization, and distinguishes recency bias from the diversity of replayed samples.

\looseness=-1Our empirical study reveals two main findings about replay in RL. First, the benefit of recency bias is strongly mediated by \volumename: when \volumename{} is small, biasing replay toward recent transitions yields substantial gains~\citep{wang2020striving}. Conversely, when \volumename{} is large, the same bias offers limited benefit. Second, \densityname{} alone does not determine performance: smoothly biased, high-entropy samplers outperform strategies that obtain comparable \densityname{} by restricting the support of the replay sampling distribution. Together, these findings explain why non-uniform replay is especially effective in optimization-limited regimes such as multi-task learning and large-scale parallel simulation, while uniform replay remains competitive when data reuse is already high.

\looseness=-1Motivated by these observations, we adopt a simple recency-biased replay sampler based on a truncated geometric distribution. The sampler directly targets the regime identified by our analysis: it increases \densityname{} smoothly while preserving high sampling entropy, requires no costly priority recomputation, and admits constant-time sampling. Across our experiments, it improves sample efficiency over uniform, priority-biased, and recency-biased replay in low-\volumename{} regimes (Figure~\ref{fig:frontpage}) while remaining effectively neutral when \volumename{} is high. This makes it a practical replacement for uniform replay in off-policy RL, especially when optimization is limited relative to data collection. We list our contributions below:

\begin{itemize}[leftmargin=10pt, itemsep=1pt, topsep=0pt]
    \item We provide a practical guide to replay sampling in modern off-policy RL through a large-scale empirical study across single-task, multi-task, and large-scale parallel training regimes. 
    
    \item We propose a per-step view of replay that separates \densityname, \volumename, and sampling entropy. This decomposition shows that recency-bias matters most when \volumename{} is low, while high sampling entropy remains important even at comparable \densityname.
    
    \item We show that a simple truncated geometric performs competitively across diverse benchmarks, improving sample efficiency over uniform and priority-based replay in low-\volumename{} regimes with negligible computational overhead, while remaining neutral in high-\volumename{} regimes.
\end{itemize}

\section{Background}

\textbf{Off-Policy RL.}
We consider infinite-horizon MDPs, where an agent interacts with an environment by selecting actions according to a policy and receives scalar rewards~\citep{puterman1994markov, sutton2018reinforcement}. We focus on off-policy deep RL methods, which improve sample efficiency by reusing data collected under previous policies~\citep{mnih2015human, fujimoto2018addressing, haarnoja2018soft}. By decoupling data collection from optimization, off-policy algorithms allow transitions to be leveraged across many gradient updates and have enabled applications including robotic control and real-world online learning during deployment~\citep{smith2022walk, seo2025learning,as2026matters}. The replay buffer enables data reuse and stabilizes learning by inducing a more slowly changing, approximately stationary training distribution~\citep{sutton2018reinforcement,mnih2015human}.

\textbf{Replay Sampling Distributions.}
Many RL methods rely on first-in-first-out (FIFO) uniform replay, which is simple and effective across a wide range of tasks~\citep{Lillicrap2015, hafner2019dream, FastTD3}. A substantial body of prior work has explored \emph{non-uniform replay sampling}, in which transitions are sampled with unequal probability~\citep{hessel2018rainbow, barth2018distributed, fujimoto2025towards}. Representative approaches include TD-error-biased Prioritized Experience Replay~\citep[PER,][]{Schaul2015} or Reliability-Adjusted Prioritized Experience Replay~\citep[ReaPER,][]{pleiss2025reliabilityadjustedprioritizedexperiencereplay}, which bias sampling based on temporal-difference error, and recency-biased Emphasizing Recent Experience~\citep[ERE,][]{wang2019boostingsoftactorcriticemphasizing}, which biases sampling toward recent transitions. While non-uniform replay has shown improvements in some settings, it often introduces additional complexity, and it remains unclear in which regimes it consistently improves over uniform replay~\citep{nauman2024bigger}.

\textbf{Recency-Biased Sampling.}
\looseness=-1
A key motivation for recency-biased replay is that newer transitions are more related to the current policy and as such may contain more informative signal, particularly when the agent adapts to new regions of the state space~\citep{wang2020striving}. Prior work analyzes this through the lens of whole-training exposure: the expected number of times a transition is sampled while it remains eligible for replay. Crucially, when the buffer size is larger than the total number of steps the agent collects over its lifetime, older transitions are sampled disproportionately more often than newer ones. ERE addresses this by skewing back the sampling distribution towards recent transitions~\citep{wang2019boostingsoftactorcriticemphasizing, wang2020striving}. While the exposure framing intuitively motivates recency-biased methods, we argue it can obscure factors that matter during optimization: two sampling strategies can share identical exposure yet differ greatly in performance, or exposure can be skewed while the choice of strategy barely matters. In contrast to prior work, we analyze replay per environment step, separating \densityname, \volumename, and sampling entropy.

\section{What Matters in Non-Uniform Sampling?}
\label{sec:analysis}

Prior work has shown that increasing replay recency can improve learning in some settings~\citep{wang2019boostingsoftactorcriticemphasizing, wang2020striving}. Our goal is therefore not to re-establish the importance of recency, but to identify the factors that determine \emph{when} recency improves over other replay sampling distributions.  

\begin{figure}
    \centering
    \includegraphics{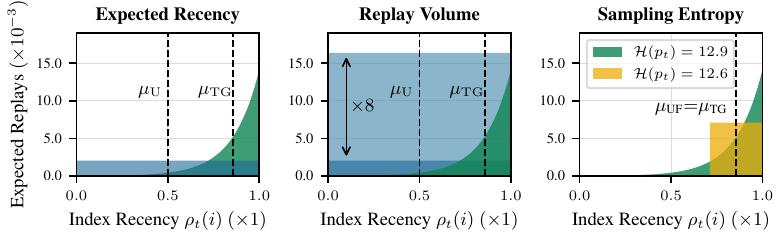}
    \vspace{-0.08in}
    \caption{
    \emph{\textbf{Effect of \densityname, \volumename, and \entropyname{}.}}
    (\textit{left}) By biasing samples toward recent data, the Truncated Geometric sampler (\Cref{sec:truncated-geometric}) achieves substantially higher \densityname{} than uniform replay.
    (\textit{middle}) Increasing \volumename{} through UTD or batch size can make uniform replay match or exceed recency-biased sampling in the number of updates applied to recent transitions.
    (\textit{right}) Replay distributions can have the same \densityname{} but different \entropyname. The Truncated Geometric sampler achieves recency through a smooth full-support bias, increasing \entropyname{} relative to a uniform buffer with the same \densityname{}; this is comparable to increasing the uniform sampling buffer size by roughly one third.
    }
    \label{fig:sampling-distributions}
    \vspace{-0.5cm}
\end{figure}

Let $p_t$ denote the replay sampling distribution at environment step $t$ over buffer indices   
$i = 0, \dots, N$, where $i = 0$ denotes the oldest sample and $N$ is the total size of the buffer. Let $N_{t}$ be the number of transitions collected up to time $t$. We define the normalized recency of index $i$ as $\rho_t(i) = i/(N_{t} - 1)$,
where larger values correspond to more recent transitions. We define the \textbf{\densityname} of the replay sampling distribution $p_t$ as
$\mu_t = \mathbb{E}_{i \sim p_t}[\rho_t(i)]$. This quantity captures how recent the sampled transitions are on average. Higher \densityname{} means that the learner updates more often on the recent experience. Unlike whole-training exposure, which aggregates counts over a transition's lifetime, \densityname{} describes the ``age'' composition of the update at a single training step.

To determine the aggressiveness of optimization at a given timestep, we define \textbf{\volumename} as $\mathrm{UTD}_t \times B_t$, that is, the number of replayed transitions between two consecutive environment steps. Increasing \volumename{} increases the amount of optimization performed on the stored data without changing the sampling distribution itself. Thus, replay can change along two separable axes: \densityname{} determines \emph{where} replay samples come from, while \volumename{} determines \emph{how many times} a given transition is consumed. Importantly, when \volumename{} is high, even low-\densityname{} samplers can still update on many recent transitions; when \volumename{} is low, the sampling distribution plays a larger role in determining whether recent transitions are replayed at all.

Furthermore, \densityname{} does not fully characterize the shape of the replay sampling distribution. Two strategies can have the same \densityname{} while differing substantially in how concentrated their sampling probabilities are. We capture this concentration using the instantaneous \textbf{sampling entropy} $\mathcal{H}(p_t)$. As illustrated in the right panel of Figure~\ref{fig:sampling-distributions}, a sampler can obtain high \densityname{} by sampling only from a recent window, yielding low entropy, or by smoothly biasing probabilities over the full buffer, yielding higher entropy. These strategies may have similar \densityname{}, but differ in update diversity and optimization dynamics.

This per-step decomposition guides the rest of our analysis. In Section~\ref{sec:sampling_scale}, we vary \text{\volumename} while keeping the replay sampling distributions fixed, showing that recency bias helps most when replay volume is low and has limited effect when replay volume is high. In Section~\ref{sec:sampling_entropy}, we compare replay strategies that bias sampling toward recent data but differ in sampling entropy, showing that recency alone is not sufficient: high-entropy, full-support sampling is also important. Together, these experiments demonstrate that separating \densityname, \volumename, and sampling entropy accounts for the observed behavior across different off-policy RL regimes.

\subsection{Does Replay Volume Matter?}
\label{sec:sampling_scale}

We first test whether \volumename{} modulates the effectiveness of recency-biased sampling.

\begin{figure}[t]
    \centering
    \includegraphics{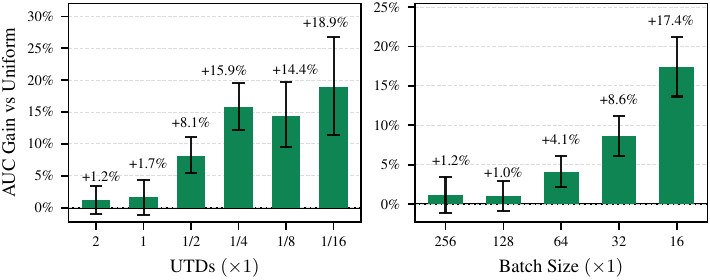}
    \vspace{-0.1in}
    \caption{
    \emph{\textbf{Replay volume matters.}}
    Improvement of recency-biased sampling over uniform replay as \volumename{} is varied through UTD (\emph{left}) and batch size (\emph{right}) (Section~\ref{sec:sampling_scale}). In both panels, reducing \volumename{} increases the advantage of recency-biased replay: when replay volume is high, both methods perform similarly, whereas when replay volume is low, recency-biased sampling yields substantially larger gains. These results show that the benefit of recency bias depends on how much replay is performed per environment step. As such, the sampling distribution is especially important in optimization-limited regimes, such as multi-task learning or large-scale parallel simulation. PER follows a similar trend (Appendix~\ref{app:per_utd_bs}).
    }
    \label{fig:utd_entropy}
    \vspace{-0.5cm}
\end{figure}

\textbf{Experiment.} \looseness=-1To isolate the role of \volumename, we vary UTD and batch size while keeping the replay buffer and replay sampling distributions fixed. Varying UTD changes the number of gradient updates performed per environment step, while varying batch size changes the number of transitions consumed per update. Both therefore change \volumename{} without changing the \densityname{} or \entropyname{} induced by a fixed sampling strategy. Both UTD and batch size are practically relevant hyperparameters and both reflect compute-data tradeoffs in modern off-policy RL systems~\citep{rybkin2025value, fu2025computeoptimalscalingvaluebaseddeep}. As such, contemporary algorithms operate across a wide range of UTD regimes and batch size settings.\footnote{E.g., BRO~\citep{nauman2024bigger} uses $\text{UTD}=10$, whereas FastTD3~\citep{FastTD3} uses $\text{UTD}=\frac{1}{64}$ due to large-scale parallel data collection.} We compare uniform replay to recency-biased Truncated Geometric sampling using SimbaV2~\citep{SimbaV2} on 13 HumanoidBench tasks~\citep{sferrazza2024humanoidbench}. We sweep UTD values $\{2, 1, \tfrac{1}{2}, \tfrac{1}{4}, \tfrac{1}{8}, \tfrac{1}{16}\}$ and batch sizes $\{256,128,64,32,16\}$. As UTD or batch size decreases, \volumename{} decreases, entering regimes where each transition is replayed relatively few times. We detail the setup in Appendix~\ref{app:experimental_details}.

\textbf{Results.} As shown in Figure~\ref{fig:utd_entropy}, uniform and recency-biased replay perform similarly at high UTD values and large batch sizes. In this regime, recency-biased replay has higher \densityname{} than uniform replay, but \volumename{} is also large, so even uniform replay updates on many recent transitions. Consequently, increasing \densityname{} provides limited additional benefit. As UTD or batch size decreases, however, the performance gap widens, with recency-biased replay increasingly outperforming uniform replay. Since the sampling strategies remain fixed across the sweep, their relative \densityname{} and \entropyname{} stay unchanged, while \volumename{} decreases. The observed shift therefore shows that the benefit of recency bias depends critically on \volumename: when \volumename{} is limited, the replay sampling distribution becomes more consequential; when \volumename{} is large, changing \densityname{} has little effect. These results show that \volumename{} modulates when non-uniform replay improves over uniform replay.

\begin{tcolorbox}[takeawaybox1]
  The benefit of increasing \densityname{} depends on \volumename. When \volumename{} is high, uniform replay already samples many recent transitions, so recency offers limited benefits. When \volumename{} is low, the sampling distribution becomes more consequential, and recency-biased replay can yield substantial gains. Thus, the replay distribution is especially important in optimization-limited regimes such as multi-task learning or large-scale parallel simulation.
\end{tcolorbox}

\newcommand{\tabFrac}{0.5}            
\newcommand{\tabFont}{\small}   
\newcommand{\tabMethodW}{2.2cm}       
\newcommand{\tabNumW}{0.9cm}          
\newcommand{\tabGainW}{1.30cm}         
\newcommand{\tabRowPad}{4pt}           
\newcommand{\tabColPad}{2pt}           
\newcommand{\tabTint}{10}               
\newcommand{\tabEreTint}{12}  

\begin{figure}[t]
    \centering
    \vspace{-0.1in}
    \begin{minipage}[c]{\tabFrac\linewidth}
    \centering
    \begin{tikzpicture}
    \matrix (m) [matrix of nodes,
        nodes={font=\tabFont, text height=1.7ex, text depth=0.5ex,
               inner xsep=\tabColPad, inner ysep=\tabRowPad,
               align=right, text width=\tabNumW},
        column 1/.style={nodes={text width=\tabMethodW, align=left}},
        column 4/.style={nodes={text width=\tabGainW}},
        column sep=0pt, row sep=0pt]
    {
      \textbf{Method}                  & $\mu$ & $\mathcal{H}(p_t)$ & \text{AUC gain} \\
      \swatch{cfresh}Trunc.\ Geom.     & 0.85 & \textbf{12.9} & $\mathbf{+12.9}\%$\\
      \swatch{c100k}Unif.\ FIFO & 0.85 & 12.6         & $+4.6\%$ \\
      \swatch{cere}ERE                 & 0.85 & 12.2  & $-41.4\%$\\
    };
    \begin{scope}[on background layer]
      \foreach \r/\c in {2/cfresh!\tabTint, 3/c100k!\tabTint, 4/cere!\tabEreTint}
        \fill[\c] (m-\r-1.north west) rectangle (m-\r-4.south east);
    \end{scope}
    \draw[thick]    (m-1-1.north west) -- (m-1-4.north east);
    \draw[black] (m-1-1.south west) -- (m-1-4.south east);
    \draw[thick]    (m-4-1.south west) -- (m-4-4.south east);
    \node[font=\small, text=black, anchor=north, yshift=-1pt]
      at (m.south) {AUC gain relative to Uniform (1M).};
    \end{tikzpicture}
    \end{minipage}%
    \hfill
    \begin{minipage}[c]{\dimexpr\linewidth-\tabFrac\linewidth-0.01\linewidth\relax}
    \centering
    \includegraphics[width=\linewidth]{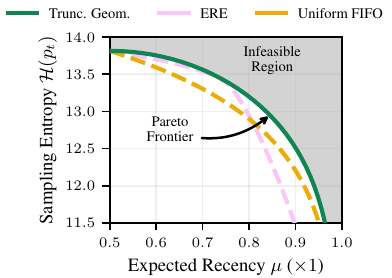}
    \end{minipage}
    \caption{\emph{\textbf{Sampling entropy matters.}}
    (\emph{left}) ERE, Uniform FIFO ($300$k), and Truncated Geometric are matched to the same \densityname{} ($\mu \approx 0.85$), yet AUC gain relative to the Uniform ($1$M) baseline differs substantially with the sampling entropy $\mathcal{H}(p_t)$. At this $\mu$, ERE has the lowest entropy and falls below the Uniform baseline (see Appendix~\ref{app:ere_shortcomings}), while Truncated Geometric, which has the highest entropy at this $\mu$, performs best. Interestingly, Uniform FIFO ($300$k) also outperforms the larger Uniform baseline ($1$M) despite using a smaller buffer. (\emph{right}) The solid curve shows the maximum-entropy frontier for a $1$M buffer: among all discrete replay sampling distributions with fixed expected recency, the Truncated Geometric distribution achieves the highest possible sampling entropy. The shaded region above the curve is infeasible. ERE and Uniform FIFO trace lower-entropy tradeoffs across hyperparameter settings, showing that they obtain recency less efficiently than Truncated Geometric; we discuss entropy interpretation in Appendix~\ref{app:entropy_scale}. }
    \label{fig:timing_frontier}
    \vspace{-0.13in}
\end{figure}

\subsection{Does Sampling Entropy Matter?}
\label{sec:sampling_entropy}

The preceding subsection shows that \volumename{} modulates the effectiveness of sampling. We now study the second factor: whether replay strategies with comparable \densityname{} can nevertheless differ in performance due to the entropy of their replay sampling distributions.


\textbf{Experiment.} 
\looseness=-1To assess whether sampling entropy matters beyond recency, we compare four replay strategies on HumanoidBench using BRC~\citep{nauman2025bigger}. Three strategies are matched for expected recency ($\mu \approx 0.85$) but differ in entropy: ERE with a $1$M buffer, Uniform FIFO with a $\approx300$K buffer, and Truncated Geometric with a $1$M buffer. Matching $\mu$ allows us to probe the effect of entropy while holding average recency fixed. As shown in Figure~\ref{fig:timing_frontier}, entropy decreases substantially from Truncated Geometric to Uniform FIFO to ERE, with ERE being the most concentrated. We also include Uniform replay with a $1$M buffer as a standard baseline. We detail the setup in  Appendix~\ref{app:experimental_details}.

\textbf{Results.} As shown in Figure~\ref{fig:timing_frontier}, performance tracks sampling entropy among the $\mu$-matched recency-biased strategies: ERE, which has the lowest entropy, underperforms even the Uniform baseline; Uniform FIFO ($300$k) performs better; and Truncated Geometric, which has the highest entropy, achieves the best performance overall. Interestingly, Uniform FIFO ($300$k) slightly outperforms Uniform ($1$M) despite using a $3$ times smaller buffer, which we attribute to its higher expected recency. These results show that \emph{how} recency is achieved matters. Strategies that obtain recency by sharply restricting sampling support sacrifice entropy of the replay sampling distribution and can underperform even when their average recency is favorable. In contrast, smooth full-support bias, as in Truncated Geometric replay, preserves update diversity while shifting replay toward recent data.

\begin{tcolorbox}[takeawaybox2]
    Expected recency is informative but not sufficient to determine performance. Strategies with comparable \densityname{} can differ substantially depending on whether they preserve high-entropy, full-support sampling. Effective non-uniform replay should therefore bias sampling smoothly while maintaining the diversity in the replay sampling distribution.
\end{tcolorbox}

\section{Efficient Truncated Geometric Sampling Procedure}
\label{sec:truncated-geometric}

The analysis in Section~\ref{sec:analysis} suggests that an effective replay sampler should increase expected recency without sacrificing sampling entropy or adding computational overhead. We instantiate this principle with a truncated geometric distribution over replay-buffer indices. This sampler smoothly biases updates toward recent transitions, admits constant-time inverse-CDF sampling, and maximizes entropy among distributions with fixed expected recency. We summarize the procedure in Algorithm~\ref{alg:recency-sampling}.

\begin{wrapfigure}[22]{r}{0.48\textwidth}
  \centering
  \vspace{-8pt}
  \includegraphics[width=0.4\textwidth]{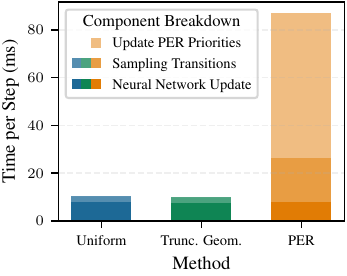}
  \vspace{-5pt}
  \caption{\looseness=-1\emph{\textbf{Decomposition of total latency.}} We compare the computational overhead of Uniform, Truncated Geometric, and PER. While the network update time is constant across methods, PER introduces significant latency due to priority tree management. In contrast, the truncated geometric sampling maintains a profile nearly identical to uniform due to efficient probability calculation.}
  \label{fig:latency}
\end{wrapfigure}

\textbf{Truncated Geometric Sampler.}
\looseness=-1
To smoothly increase \densityname{} while retaining constant-time sampling, we use a truncated geometric distribution over replay-buffer indices. Let $N_{\max}$ denote the maximum replay-buffer capacity and $N_{\text{current}}$ its size at optimization step $t$. Transitions are indexed from oldest ($i=0$) to newest ($i=N_{\text{current}}-1$). We sample an index $i \in \{0,\dots,N_{\text{current}}-1\}$ according to
\[
    p(i) \;\propto\; 2^{\,\alpha \frac{i}{N_{\max}-1}},
\]
where $\alpha > 0$ controls the strength of the recency bias. Larger values of $\alpha$ place more probability mass on recent transitions, while $\alpha \to 0$ recovers uniform replay. Unlike PER, we omit importance-sampling correction, which would counteract the recency bias. This sampler directly addresses the factors identified in Section~\ref{sec:analysis}. Firstly, it admits closed-form inverse-CDF sampling and therefore constant-time implementation. Secondly, in contrast to approaches like ERE, it increases \densityname{} smoothly, without restricting support to a recent window. Finally, among all distributions over buffer indices with fixed \densityname{}, the truncated geometric distribution maximizes \entropyname~\citep{Jaynes1957, PMEGeom}. Thus, for a given expected recency, it preserves the highest update diversity. We visualize the tradeoff between \densityname{} and \entropyname{} in the right panel of Figure~\ref{fig:timing_frontier}.

\textbf{Efficient Sampling.} 
\looseness=-1
Sampling is performed using an efficient inverse transform. A single uniform random variable $U \sim \mathrm{Unif}(0,1)$ is mapped through the closed-form inverse cumulative distribution function of the truncated geometric distribution. The sampled index is given by:
\[
i =  
\left\lfloor\frac{\log_2\!\big(1 + U \cdot Z \cdot (2^k - 1)\big)}{k}\right\rfloor
- 1
\]
clipped to $[0, N_{\text{current}}-1]$, with $k = \alpha \mathbin{/} (N_{\max}-1)$, and
$Z = (2^{k N_{\text{current}}}-1) \mathbin{/} (2^k-1)$. While non-uniform replay is sometimes assumed to introduce additional wall-clock overhead, the proposed sampling procedure requires constant time, yielding $\mathcal{O}(1)$ complexity. This contrasts with the cost incurred by strategies whose sampling probabilities depend on updated priorities supported by segment trees \citep{Schaul2015}, which yield $\mathcal{O}(\log N)$ complexity. As shown in Figure~\ref{fig:latency}, the empirical wall-clock performance of the proposed truncated geometric sampling is comparable to uniform replay.

 \textbf{Interpretation of $\boldsymbol{\alpha}$.} The proposed truncated geometric sampling is governed by a single hyperparameter $\alpha$. As we now show, it provides intuitive control over \textbf{\densityname} in sampling. Approximately, for
$\Delta \approx (N_{\max}-1)/\alpha$,
\[
\frac{p(i + \Delta)}{p(i)} \approx 2.
\]
Thus, moving forward by roughly $(N_{\max}-1)/\alpha$ positions doubles the sampling probability. Smaller $\alpha$ values approach uniform replay, while larger values increasingly emphasize recent transitions. We discuss how we choose $\alpha$ in Appendix~\ref{app:alpha_choice}.

\begin{algorithm}[t!]
\caption{Efficient Truncated Geometric Sampling}
\begin{algorithmic}[1]
\REQUIRE Replay buffer $\mathcal{D}$, capacity $N_{\max}$, current size $N_{\text{current}} \le N_{\max}$, recency parameter $\alpha>0$.
\vspace{-0.2em}
\item[] \hrulefill
\STATE Index transitions in $\mathcal{D}$ as $i=0,\dots,N_{\text{current}}-1$, where $i=0$ is the oldest and $i=N_{\text{current}}-1$ is the most recent.
\STATE Set $k \leftarrow \alpha / (N_{\max}-1)$.
\STATE Sample $U \sim \mathrm{Unif}(0,1)$.
\STATE Compute normalization constant $Z \leftarrow \smash{\frac{2^{k N_{\text{current}}}-1}{2^{k}-1}}$.
\STATE Compute $r \leftarrow U \cdot Z$.
\STATE Compute index via inverse CDF: $i \leftarrow \lfloor \smash{\frac{\log_2(1 + r(2^k-1))}{k} - 1 \rfloor}$.
\STATE Clip $i$ to $[0, N_{\text{current}}-1]$.
\STATE \textbf{return} transition $\mathcal{D}[i]$.
\end{algorithmic}
\label{alg:recency-sampling}
\end{algorithm}

\section{Experiments}
\label{sec:experiments}

The analysis in Section~\ref{sec:analysis} predicts that recency-biased sampling should be most effective when \volumename{} is limited, provided the sampler preserves high entropy. We test this prediction across four representative off-policy protocols: large-scale parallel simulation, multi-task learning, single-task learning, and discrete-action tasks. Parallel simulation and multi-task learning are the primary low-\volumename{} regimes, since data collection is distributed across many environments or tasks.

For each regime, we start from the original algorithm and training setup of the corresponding source paper, modifying only the replay sampler. We compare the original uniform replay baseline against PER~\citep{Schaul2015}, ERE~\citep{wang2019boostingsoftactorcriticemphasizing}, and the proposed Truncated Geometric sampler. All base algorithm hyperparameters and training protocols are kept unchanged. PER and ERE hyperparameters are tuned following Appendix~\ref{app:hyperparams}, while Truncated Geometric uses a fixed $\alpha=10$ across all experiments. Unless otherwise specified, all methods use a replay buffer of capacity $1$M. We report $95\%$ stratified bootstrap CIs computed with RLiable~\citep{RLiable}. Performance is evaluated using normalized episodic returns and success rates where applicable, across multiple seeds. Full experimental details, including hyperparameter settings, detailed figure descriptions, computational resources, environment lists, normalization procedures, and evaluation protocols are provided in Appendices~\ref{app:experimental_details} and~\ref{app:training_details}.

\begin{figure}[t!]
    \centering
    \vspace{-0.2in}    \includegraphics{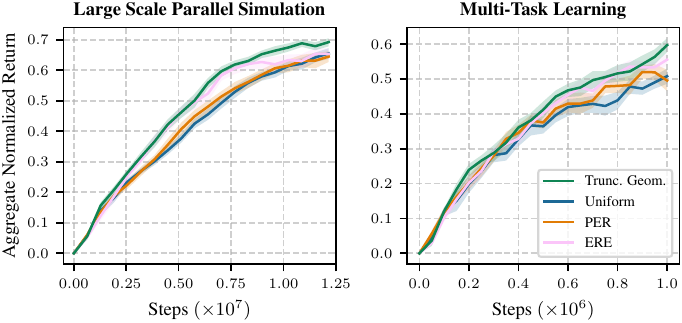}
\caption{
    \emph{\textbf{High-dimensional humanoid locomotion and manipulation tasks.}}
    We report aggregate mean return across 29 HumanoidBench tasks trained in a parallel simulation setup with FastTD3 (\textit{left}) and 20 HumanoidBench tasks trained in a multi-task setup with BRC (\textit{right}). Shaded regions show $95\%$ CIs. Truncated Geometric sampling significantly improves over uniform replay on both benchmarks and outperforms PER and ERE, despite FastTD3 and BRC already being strong optimized baselines. These gains come with negligible computational and complexity overhead relative to uniform sampling. We detail the experimental setup in Section~\ref{sec:experiments} and Appendix~\ref{app:experimental_details}.
    }
    \label{fig:learning_curve}
    \vspace{-0.2in}
\end{figure}

\textbf{Large-Scale Parallel Simulation.} We first evaluate replay sampling in a large-scale parallel simulation setting using FastTD3~\citep{FastTD3}. We consider 29 high-dimensional continuous-control tasks from HumanoidBench~\citep{sferrazza2024humanoidbench} and 6 tasks from Isaac Lab~\citep{isaaclab}; full task lists are provided in Appendix~\ref{app:task_lists}. In this setting, experience is collected rapidly from many environments in parallel, while relatively few gradient updates are performed per environment step, resulting in limited \volumename{}. On HumanoidBench, shown in the left panel of Figure~\ref{fig:learning_curve}, Truncated Geometric sampling improves sample efficiency over uniform replay, achieving up to $1.35\times$ faster learning in terms of environment interactions. PER performs comparably to uniform replay, while ERE provides intermediate gains between uniform replay and Truncated Geometric. Because Truncated Geometric sampling incurs small computational overhead relative to uniform, these sample-efficiency gains translate directly into faster wall-clock convergence, reaching comparable performance in approximately $80\%$ of the training time required by standard FastTD3 with uniform replay. We observe similar trends on Isaac Lab, where recency-biased samplers also outperform uniform replay. In this benchmark, Truncated Geometric performs comparably to ERE in aggregate, with ERE benefiting from particularly strong performance on one task. We report the full Isaac Lab results in Appendix~\ref{app:fasttd3-isaac}.

\textbf{Multi-Task Learning.} 
\looseness=-1
We next evaluate replay sampling in the multi-task setting using BRC~\citep{nauman2025bigger} on the HumanoidBench-Hard tasks~\citep{nauman2025bigger} listed in Appendix~\ref{app:multi-task-benchmarks}. As in large-scale parallel simulation, multi-task learning naturally limits \volumename{}, since experience is distributed across many tasks while optimization per task remains limited. As shown in the right panel of Figure~\ref{fig:learning_curve}, Truncated Geometric sampling achieves the best overall performance and improves substantially over uniform replay. ERE is competitive in this setting, though slightly below Truncated Geometric, in contrast to the low-entropy ERE configurations studied in Section~\ref{sec:sampling_entropy}, where aggressive concentration hurts performance. PER provides modest gains over uniform replay but incurs substantial computational overhead. These results further support the analysis in Section~\ref{sec:analysis}: when replay volume is limited relative to data collection, increasing \densityname{} through smooth, high-entropy recency bias improves performance while preserving computational efficiency. We also evaluate BRC with different replay strategies on DMC Dog~\citep{DMC} and Meta-World~\citep{metaworld}; full results are provided in Appendix~\ref{app:results}.

\textbf{Single-Task Learning.}
In Section~\ref{sec:sampling_scale}, we evaluated SimbaV2 on 13 HumanoidBench Nohands tasks across UTD values from $2$ to $1/16$ and batch sizes from $256$ to $16$, showing that Truncated Geometric sampling yields larger gains as \volumename{} becomes limited. Here, we return to the original SimbaV2 benchmark of 30 tasks under its standard high-UTD, standard-batch-size setting, where \volumename{} is substantial~\citep{SimbaV2}. This experiment tests whether Truncated Geometric sampling remains safe in the high-\volumename{} regime, where our analysis predicts limited gains from increasing \densityname{}. As shown in Figure~\ref{fig:simba-learningcurve}, Truncated Geometric sampling performs on par with uniform replay, yielding a modest $1.2\%$ improvement. These results support the prediction from Section~\ref{sec:analysis}: recency-biased replay improves performance when \volumename{} is limited, while remaining effectively neutral when replay volume is high.

\paragraph{Discrete Actions and Pixel Observations.}
To test whether our findings extend beyond continuous control, we evaluate
Truncated Geometric sampling with DQN~\citep{mnih2015human}, using the CleanRL
implementation~\citep{huang2022cleanrl}, on 56 Atari games~\citep{bellemare2013arcade}.
This setting combines pixel observations, discrete actions, and low replay volume
(UTD $=1/4$, batch size $32$). Keeping all DQN hyperparameters unchanged and using
the same $\alpha = 10$, Truncated Geometric improves the AUC of the IQM
human-normalized score, the
standard Atari metric~\citep{RLiable,mnih2015human}, by $18.9\%$ over uniform replay, consistent with our prediction that recency
bias helps most when replay volume is low.
Full results are in Appendix~\ref{app:atari}.

\begin{figure}[t!]
    \centering
    \vspace{-0.1in}
    \includegraphics[width=0.95\textwidth]{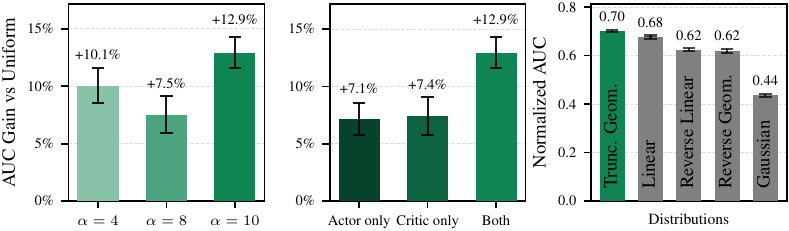}
    \vspace{-0.05in}
    \caption{
    \emph{\textbf{Ablations on Truncated Geometric replay.}} (\emph{left}) Sample efficiency gains for Truncated Geometric replay with different recency parameter $\alpha$, using $20$ HumanoidBench tasks. Truncated Geometric consistently improves over uniform replay across all tested values of $\alpha$, demonstrating robustness to hyperparameter selection. (\emph{middle}) Sample efficiency gains when Truncated Geometric replay is applied separately to actor or critic, using $20$ HumanoidBench tasks. These results show that Truncated Geometric sampling benefits actor and critic roughly equally. (\emph{right}) Sampling distributions' sample efficiency investigated on the $4$ DMC dog tasks, with Truncated Geometric as the top-performing method.}
    \label{fig:ablations}
    \vspace{-0.21in}
\end{figure}

\textbf{Ablation Studies.}
We ablate key design choices of Truncated Geometric sampling in Figure~\ref{fig:ablations} using BRC on a variety of multi-task benchmarks. First, we evaluate sensitivity to the recency parameter $\alpha$ described in Section~\ref{sec:truncated-geometric}. Truncated Geometric sampling improves over uniform replay across a wide range of values, with gains between $7.5\%$ and $12.9\%$, indicating that the Truncated Geometric sampling is robust to its only hyperparameter. Second, we ask whether the gains from recency-biased replay arise primarily through critic learning or through policy updates. This distinction is important because prior work has often found critic learning to be especially sensitive to algorithmic design choices in off-policy RL~\citep{d2022sample,nauman2024bigger}. We therefore apply Truncated Geometric sampling separately to the actor and critic updates, allowing for updates on independent batches of data. Shapley value analysis attributes $6.6\%$ of the improvement to the critic and $6.3\%$ to the actor, suggesting that both components benefit comparably from recency-biased replay. Finally, we compare against alternative non-uniform sampling distributions. Among the evaluated strategies that we describe in Appendix~\ref{app:distributions}, smooth recency-biased distributions achieve the strongest performance, supporting the role of high-entropy recency bias. We describe the experimental settings in detail in Appendix~\ref{app:hyperparams}.

\begin{tcolorbox}[takeawaybox3]
    Recency-biased replay provides the largest gains in low-\volumename{} regimes, such as large-scale parallel simulation and multi-task learning, while its benefits diminish when \volumename{} is high, as in high-UTD single-task learning. The results also support the role of sampling entropy: smooth, full-support recency bias outperforms strategies such as ERE that increase expected recency by sharply restricting sampling support. The same analysis explains Truncated Geometric sampling: it improves sample efficiency in low-\volumename{} regimes, incurs negligible overhead, and preserves performance when \volumename{} is high. Together, these findings support the analysis in Section~\ref{sec:analysis} and clarify when recency-biased replay improves over uniform replay.
\end{tcolorbox}

\section{Conclusions}

\looseness=-1We revisited non-uniform replay sampling in off-policy reinforcement learning and showed that its effectiveness cannot be understood from expected recency alone. Instead, replay performance is mediated by two additional factors: \volumename{} and \entropyname. Recency-biased replay is most beneficial in low-\volumename{} regimes, where transitions are replayed infrequently, and when the sampling distribution preserves high entropy rather than sharply restricting support. Practically, these findings suggest that replay design should be treated as a function of the learner--collector ratio: uniform replay is often sufficient when optimization is abundant, while smooth recency-biased replay is a simple and effective default when data collection outpaces learning.

Guided by this analysis, we introduced a simple Truncated Geometric replay sampler that smoothly increases expected recency while maintaining high entropy and constant-time sampling. Across large-scale parallel simulation, multi-task learning, and single-task settings, this approach improves sample efficiency in low-\volumename{} regimes, incurs negligible computational overhead with $\mathcal{O}(1)$ sampling complexity, and remains neutral when \volumename{} is high. Our results clarify when and why recency-biased replay improves learning, and provide a practical alternative to uniform and priority-based replay for modern off-policy RL. We discuss limitations of our analysis in Appendix~\ref{app:limitations}.

\section*{Acknowledgments}
\looseness=-1We gratefully acknowledge Polish high-performance computing infrastructure PLGrid (HPC Center: ACK Cyfronet AGH) for providing computer facilities and support within computational grant no. PLG/2025/018697. Mikołaj Czarnecki was supported by the National Science Centre under Grant No. 2023/50/E/ST6/00665.
Yarden As received funding from a grant of the Hasler foundation~(grant no. 21039) and the ETH AI Center.
Pieter Abbeel holds concurrent appointments as a Professor
at UC Berkeley and as an Amazon Scholar. This paper
describes work performed at UC Berkeley and is not
associated with Amazon.

\newpage

\bibliography{neurips_2026}

@misc{wang2019boostingsoftactorcriticemphasizing,
      title={Boosting Soft Actor-Critic: Emphasizing Recent Experience without Forgetting the Past}, 
      author={Che Wang and Keith Ross},
      year={2019},
      eprint={1906.04009},
      archivePrefix={arXiv},
      primaryClass={cs.LG},
      url={https://arxiv.org/abs/1906.04009}, 
}

@article{nauman2024bigger,
  title={Bigger, regularized, optimistic: scaling for compute and sample efficient continuous control},
  author={Nauman, Michal and Ostaszewski, Mateusz and Jankowski, Krzysztof and Mi{\l}o{\'s}, Piotr and Cygan, Marek},
  journal={Advances in neural information processing systems},
  year={2024}
}

@inproceedings{haarnoja2018soft,
  title={Soft actor-critic: Off-policy maximum entropy deep reinforcement learning with a stochastic actor},
  author={Haarnoja, Tuomas and Zhou, Aurick and Abbeel, Pieter and Levine, Sergey},
  booktitle={International conference on machine learning},
  year={2018},
}

@inproceedings{fujimoto2018addressing,
  title={Addressing function approximation error in actor-critic methods},
  author={Fujimoto, Scott and Hoof, Herke and Meger, David},
  booktitle={International conference on machine learning},
  year={2018},
}

@article{Jaynes1957,
  title={Information theory and statistical mechanics},
  author={Jaynes, Edwin T},
  journal={Physical review},
  volume={106},
  number={4},
  pages={620},
  year={1957},
  publisher={APS}
}

@book{PMEGeom,
  title={Elements of information theory},
  author={Cover, Thomas M},
  year={1999},
  publisher={John Wiley \& Sons}
}

@article{FastTD3,
  title={FastTD3: Simple, Fast, and Capable Reinforcement Learning for Humanoid Control},
  author={Seo, Younggyo and Sferrazza, Carmelo and Geng, Haoran and Nauman, Michal and Yin, Zhao-Heng and Abbeel, Pieter},
  journal={arXiv preprint arXiv:2505.22642},
  year={2025}
}

@article{seo2025learning,
  title={Learning Sim-to-Real Humanoid Locomotion in 15 Minutes},
  author={Seo, Younggyo and Sferrazza, Carmelo and Chen, Juyue and Shi, Guanya and Duan, Rocky and Abbeel, Pieter},
  journal={arXiv preprint arXiv:2512.01996},
  year={2025}
}

@article{SimbaV2,
  title={Hyperspherical normalization for scalable deep reinforcement learning},
  author={Lee, Hojoon and Lee, Youngdo and Seno, Takuma and Kim, Donghu and Stone, Peter and Choo, Jaegul},
  journal={arXiv preprint arXiv:2502.15280},
  year={2025}
}

@article{RLiable,
  title={Deep Reinforcement Learning at the Edge of the Statistical Precipice},
  author={Agarwal, Rishabh and Schwarzer, Max and Castro, Pablo Samuel
          and Courville, Aaron and Bellemare, Marc G},
  journal={Advances in Neural Information Processing Systems},
  year={2021}
}

@article{DMC,
  title={Deepmind control suite},
  author={Tassa, Yuval and Doron, Yotam and Muldal, Alistair and Erez, Tom and Li, Yazhe and Casas, Diego de Las and Budden, David and Abdolmaleki, Abbas and Merel, Josh and Lefrancq, Andrew and others},
  journal={arXiv preprint arXiv:1801.00690},
  year={2018}
}

@article{sferrazza2024humanoidbench,
  title={Humanoidbench: Simulated humanoid benchmark for whole-body locomotion and manipulation},
  author={Sferrazza, Carmelo and Huang, Dun-Ming and Lin, Xingyu and Lee, Youngwoon and Abbeel, Pieter},
  journal={arXiv preprint arXiv:2403.10506},
  year={2024}
}

@article{nauman2025bigger,
  title={Bigger, Regularized, Categorical: High-Capacity Value Functions are Efficient Multi-Task Learners},
  author={Nauman, Michal and Cygan, Marek and Sferrazza, Carmelo and Kumar, Aviral and Abbeel, Pieter},
  journal={arXiv preprint arXiv:2505.23150},
  year={2025}
}

@inproceedings{barth2018distributed,
  title={Distributed Distributional Deterministic Policy Gradients},
  author={Barth-Maron, Gabriel and Hoffman, Matthew W and Budden, David and Dabney, Will and Horgan, Dan and Dhruva, TB and Muldal, Alistair and Heess, Nicolas and Lillicrap, Timothy},
  booktitle={International Conference on Learning Representations},
  year={2018}
}

@inproceedings{wang2020striving,
  title={Striving for simplicity and performance in off-policy DRL: Output normalization and non-uniform sampling},
  author={Wang, Che and Wu, Yanqiu and Vuong, Quan and Ross, Keith},
  booktitle={International Conference on Machine Learning},
  pages={10070--10080},
  year={2020},
  organization={PMLR}
}

@article{smith2022walk,
  title={A walk in the park: Learning to walk in 20 minutes with model-free reinforcement learning},
  author={Smith, Laura and Kostrikov, Ilya and Levine, Sergey},
  journal={arXiv preprint arXiv:2208.07860},
  year={2022}
}

@inproceedings{d2022sample,
  title={Sample-Efficient Reinforcement Learning by Breaking the Replay Ratio Barrier},
  author={D'Oro, Pierluca and Schwarzer, Max and Nikishin, Evgenii and Bacon, Pierre-Luc and Bellemare, Marc G and Courville, Aaron},
  booktitle={The Eleventh International Conference on Learning Representations},
  year={2022}
}

@article{silver2017mastering,
  title={Mastering the game of go without human knowledge},
  author={Silver, David and Schrittwieser, Julian and Simonyan, Karen and Antonoglou, Ioannis and Huang, Aja and Guez, Arthur and Hubert, Thomas and Baker, Lucas and Lai, Matthew and Bolton, Adrian and others},
  journal={nature},
  volume={550},
  number={7676},
  pages={354--359},
  year={2017},
  publisher={Nature Publishing Group}
}

@inproceedings{hafner2019dream,
  title={Dream to Control: Learning Behaviors by Latent Imagination},
  author={Hafner, Danijar and Lillicrap, Timothy and Ba, Jimmy and Norouzi, Mohammad},
  booktitle={International Conference on Learning Representations},
  year={2019}
}

@article{huang2022cleanrl,
  author  = {Shengyi Huang and Rousslan Fernand Julien Dossa and Chang Ye and Jeff Braga and Dipam Chakraborty and Kinal Mehta and João G.M. Araújo},
  title   = {CleanRL: High-quality Single-file Implementations of Deep Reinforcement Learning Algorithms},
  journal = {Journal of Machine Learning Research},
  year    = {2022},
  volume  = {23},
  number  = {274},
  pages   = {1--18},
}

@article{bellemare2013arcade,
  title={The arcade learning environment: An evaluation platform for general agents},
  author={Bellemare, Marc G and Naddaf, Yavar and Veness, Joel and Bowling, Michael},
  journal={Journal of Artificial Intelligence Research},
  volume={47},
  pages={253--279},
  year={2013}
}

@article{caggiano2022myosuite,
  title={MyoSuite--A contact-rich simulation suite for musculoskeletal motor control},
  author={Caggiano, Vittorio and Wang, Huawei and Durandau, Guillaume and Sartori, Massimo and Kumar, Vikash},
  journal={arXiv preprint arXiv:2205.13600},
  year={2022}
}

@STRING{iclr = {ICLR}}

@STRING{aaai = {AAAI}}

@inproceedings{bhatt2024crossq,
    title        = {Cross{Q}: Batch Normalization in Deep Reinforcement Learning for Greater Sample Efficiency and Simplicity},
    author       = {Aditya Bhatt and Daniel Palenicek and Boris Belousov and Max Argus and Artemij Amiranashvili and Thomas Brox and Jan Peters},
    year         = 2024,
    booktitle    = {International conference on learning representations (ICLR)}
}

@article{harris2020array,
  title={Array programming with NumPy},
  author={Harris, Charles R and Millman, K Jarrod and Van Der Walt, St{\'e}fan J and Gommers, Ralf and Virtanen, Pauli and Cournapeau, David and Wieser, Eric and Taylor, Julian and Berg, Sebastian and Smith, Nathaniel J and others},
  journal={Nature},
  volume={585},
  number={7825},
  pages={357--362},
  year={2020},
  publisher={Nature Publishing Group UK London}
}

@book{puterman1994markov,
  title={Markov Decision Processes: Discrete Stochastic Dynamic Programming},
  author={Puterman, Martin L},
  year={1994},
  publisher={John Wiley \& Sons, Inc.}
}

@article{lillicrap2015,
  title={Continuous control with deep reinforcement learning},
  author={Lillicrap, Timothy P and Hunt, Jonathan J and Pritzel, Alexander and Heess, Nicolas and Erez, Tom and Tassa, Yuval and Silver, David and Wierstra, Daan},
  journal={arXiv preprint arXiv:1509.02971},
  year={2015}
}

@book{sutton2018reinforcement,
  title={Reinforcement learning: An introduction},
  author={Sutton, Richard S and Barto, Andrew G},
  year={2018},
  publisher={MIT press}
}

@inproceedings{hessel2018rainbow,
  title={Rainbow: Combining improvements in deep reinforcement learning},
  author={Hessel, Matteo and Modayil, Joseph and Van Hasselt, Hado and Schaul, Tom and Ostrovski, Georg and Dabney, Will and Horgan, Dan and Piot, Bilal and Azar, Mohammad and Silver, David},
  booktitle={Thirty-Second AAAI Conference on Artificial Intelligence},
  year={2018}
}

@article{Schaul2015,
  title={Prioritized experience replay},
  author={Schaul, Tom and Quan, John and Antonoglou, Ioannis and Silver, David},
  journal={arXiv preprint arXiv:1511.05952},
  year={2015}
}

@article{rybkin2025value,
  title={Value-Based Deep RL Scales Predictably},
  author={Rybkin, Oleh and Nauman, Michal and Fu, Preston and Snell, Charlie and Abbeel, Pieter and Levine, Sergey and Kumar, Aviral},
  journal={arXiv preprint arXiv:2502.04327},
  year={2025}
}

@article{palenicek2025xqc,
  title={XQC: Well-conditioned Optimization Accelerates Deep Reinforcement Learning},
  author={Palenicek, Daniel and Vogt, Florian and Watson, Joe and Posner, Ingmar and Peters, Jan},
  journal={arXiv preprint arXiv:2509.25174},
  year={2025}
}

@article{fujimoto2025towards,
  title={Towards general-purpose model-free reinforcement learning},
  author={Fujimoto, Scott and D'Oro, Pierluca and Zhang, Amy and Tian, Yuandong and Rabbat, Michael},
  journal={arXiv preprint arXiv:2501.16142},
  year={2025}
}

@article{metaworld,
  author       = {Tianhe Yu and
                  Deirdre Quillen and
                  Zhanpeng He and
                  Ryan Julian and
                  Karol Hausman and
                  Chelsea Finn and
                  Sergey Levine},
  title        = {Meta-World: {A} Benchmark and Evaluation for Multi-Task and Meta Reinforcement
                  Learning},
  journal      = {CoRR},
  volume       = {abs/1910.10897},
  year         = {2019},
  url          = {http://arxiv.org/abs/1910.10897},
  eprinttype   = {arXiv},
  eprint       = {1910.10897},
  bibsource    = {dblp computer science bibliography, https://dblp.org}
}

@article{isaaclab,
   title={Isaac Lab: A GPU-Accelerated Simulation Framework for Multi-Modal Robot Learning},
   author={Mayank Mittal and Pascal Roth and James Tigue and Antoine Richard and Octi Zhang and Peter Du and Antonio Serrano-Muñoz and Xinjie Yao and René Zurbrügg and Nikita Rudin and Lukasz Wawrzyniak and Milad Rakhsha and Alain Denzler and Eric Heiden and Ales Borovicka and Ossama Ahmed and Iretiayo Akinola and Abrar Anwar and Mark T. Carlson and Ji Yuan Feng and Animesh Garg and Renato Gasoto and Lionel Gulich and Yijie Guo and M. Gussert and Alex Hansen and Mihir Kulkarni and Chenran Li and Wei Liu and Viktor Makoviychuk and Grzegorz Malczyk and Hammad Mazhar and Masoud Moghani and Adithyavairavan Murali and Michael Noseworthy and Alexander Poddubny and Nathan Ratliff and Welf Rehberg and Clemens Schwarke and Ritvik Singh and James Latham Smith and Bingjie Tang and Ruchik Thaker and Matthew Trepte and Karl Van Wyk and Fangzhou Yu and Alex Millane and Vikram Ramasamy and Remo Steiner and Sangeeta Subramanian and Clemens Volk and CY Chen and Neel Jawale and Ashwin Varghese Kuruttukulam and Michael A. Lin and Ajay Mandlekar and Karsten Patzwaldt and John Welsh and Huihua Zhao and Fatima Anes and Jean-Francois Lafleche and Nicolas Moënne-Loccoz and Soowan Park and Rob Stepinski and Dirk Van Gelder and Chris Amevor and Jan Carius and Jumyung Chang and Anka He Chen and Pablo de Heras Ciechomski and Gilles Daviet and Mohammad Mohajerani and Julia von Muralt and Viktor Reutskyy and Michael Sauter and Simon Schirm and Eric L. Shi and Pierre Terdiman and Kenny Vilella and Tobias Widmer and Gordon Yeoman and Tiffany Chen and Sergey Grizan and Cathy Li and Lotus Li and Connor Smith and Rafael Wiltz and Kostas Alexis and Yan Chang and David Chu and Linxi "Jim" Fan and Farbod Farshidian and Ankur Handa and Spencer Huang and Marco Hutter and Yashraj Narang and Soha Pouya and Shiwei Sheng and Yuke Zhu and Miles Macklin and Adam Moravanszky and Philipp Reist and Yunrong Guo and David Hoeller and Gavriel State},
   journal={arXiv preprint arXiv:2511.04831},
   year={2025},
   url={https://arxiv.org/abs/2511.04831}
}

@misc{fu2025computeoptimalscalingvaluebaseddeep,
      title={Compute-Optimal Scaling for Value-Based Deep RL}, 
      author={Preston Fu and Oleh Rybkin and Zhiyuan Zhou and Michal Nauman and Pieter Abbeel and Sergey Levine and Aviral Kumar},
      year={2025},
      eprint={2508.14881},
      archivePrefix={arXiv},
      primaryClass={cs.LG},
      url={https://arxiv.org/abs/2508.14881}, 
}

@article{as2026matters,
  title={What Matters for Simulation to Online Reinforcement Learning on Real Robots},
  author={As, Yarden and Tirumala, Dhruva and Zurbr{\"u}gg, Ren{\'e} and Li, Chenhao and Coros, Stelian and Krause, Andreas and Wulfmeier, Markus},
  journal={arXiv preprint arXiv:2602.20220},
  year={2026}
}

@misc{pleiss2025reliabilityadjustedprioritizedexperiencereplay,
      title={Reliability-Adjusted Prioritized Experience Replay}, 
      author={Leonard S. Pleiss and Tobias Sutter and Maximilian Schiffer},
      year={2025},
      eprint={2506.18482},
      archivePrefix={arXiv},
      primaryClass={cs.LG},
      url={https://arxiv.org/abs/2506.18482}, 
}

@article{mnih2015human,
  title={Human-level control through deep reinforcement learning},
  author={Mnih, Volodymyr and Kavukcuoglu, Koray and Silver, David and Rusu, Andrei A and Veness, Joel and Bellemare, Marc G and Graves, Alex and Riedmiller, Martin and Fidjeland, Andreas K and Ostrovski, Georg and others},
  journal={nature},
  volume={518},
  number={7540},
  pages={529--533},
  year={2015},
  publisher={Nature Publishing Group UK, London}
}
\bibliographystyle{neurips_2026}

\newpage
\appendix
\section{Limitations}
\label{app:limitations}
Our experiments focus primarily on continuous-control off-policy RL, where replay buffers are central and low-\volumename{} regimes arise naturally in large-scale parallel and multi-task training. Although our ablations show that Truncated Geometric replay is robust to the recency parameter, adaptive schedules for recency bias may further improve performance. More broadly, our study is empirical: the decomposition into \densityname, \volumename, and \entropyname{} provides an explanatory and practical framework, but not a complete theory of replay. Developing such a theory could clarify how replay sampling interacts with optimization, bootstrapping, distribution shift, and function approximation.  Finally, while our analysis aligns well with aggregate trends across replay regimes, it does not fully account for task-level heterogeneity within a fixed regime. Identifying the environment or optimization properties that determine when recency-biased replay underperforms remains an important direction for future work.

\section{Truncated Geometric Sampling Procedure Details}
\label{app:truncated_geom}
\looseness=-1In this section, we discuss the implementation details of the proposed Truncated Geometric Sampling Procedure.

\subsection{Python Implementation}
We provide a concise NumPy~\citep{harris2020array} implementation of the Truncated Geometric sampling. The implementation uses inverse-CDF sampling, making it $\mathcal{O}(1)$ and thus as computationally efficient as uniform sampling.

\newcommand{\pykw}[1]{\textcolor[rgb]{0.00,0.50,0.00}{\textbf{#1}}}
\newcommand{\pynn}[1]{\textcolor[rgb]{0.00,0.00,1.00}{\textbf{#1}}}
\newcommand{\pynf}[1]{\textcolor[rgb]{0.00,0.00,1.00}{#1}}
\newcommand{\pynum}[1]{\textcolor[rgb]{0.40,0.40,0.40}{#1}}
\newcommand{\pyop}[1]{\textcolor[rgb]{0.40,0.40,0.40}{#1}}

\DefineVerbatimEnvironment{pyblock}{Verbatim}{
  commandchars=\\\{\},
  fontsize=\small
}

\begin{pyblock}
\pykw{import} \pynn{numpy} \pykw{as} \pynn{np}

\pykw{def} \pynf{geometric\_sample}(size, batch_size, alpha\pyop{=}\pynum{10.0}, capacity\pyop{=}\pynum{1\_000\_000}, offset\pyop{=}\pynum{0}):
    k \pyop{=} alpha \pyop{/} (capacity \pyop{-} \pynum{1})
    base \pyop{=} \pynum{2.0} \pyop{**} k
    Z \pyop{=} (base \pyop{**} size \pyop{-} \pynum{1.0}) \pyop{/} (base \pyop{-} \pynum{1.0})
    r \pyop{=} np\pyop{.}random\pyop{.}uniform(\pynum{0.0}, Z, size\pyop{=}batch_size)
    i \pyop{=} np\pyop{.}log2(\pynum{1.0} \pyop{+} r \pyop{*} (base \pyop{-} \pynum{1.0})) \pyop{/} k \pyop{-} \pynum{1.0}
    i \pyop{=} np\pyop{.}clip(np\pyop{.}floor(i)\pyop{.}astype(np\pyop{.}int64), \pynum{0}, size \pyop{-} \pynum{1})
    \pykw{if} offset \pyop{!=} \pynum{0}:
        i \pyop{=} (i \pyop{+} offset) \pyop{\%} capacity
    \pykw{return} i
    \end{pyblock}

\subsection{Parallel Buffer Pseudocode}
Algorithm~\ref{alg:recency-sampling-parallel} extends the single-task Truncated Geometric sampling procedure to the multi-task setting, where each task maintains an independent replay buffer. At each sampling step, a task is selected uniformly at random, and a transition is sampled from the corresponding buffer using the Truncated Geometric distribution, preserving recency bias across all tasks without introducing additional computational overhead.

\begin{algorithm}[ht!]
\caption{Efficient Truncated Geometric Sampling Procedure}
\begin{algorithmic}[1]
\REQUIRE Per-task replay buffer $\mathcal{D}$, max capacity $N_{\max}$, current size $N_{\text{current}} \le N_{\max}$, number of tasks $M$, recency parameter $\alpha>0$.
\vspace{-0.6em}
\item[]  \hrulefill
\STATE Sample task $j \sim \mathrm{Unif}(\{0,\dots,M-1\})$.
\STATE Index transitions in $\mathcal{D}[j]$ as $i=0,\dots,N_{\text{current}}-1$, where $i=0$ is the oldest and $i=N_{\text{current}}-1$ is the most recent.
\STATE Set $k \leftarrow \alpha / (N_{\max}-1)$.
\STATE Sample $U \sim \mathrm{Unif}(0,1)$.
\STATE Compute normalization constant $Z \leftarrow \smash{\frac{2^{k N_{\text{current}}}-1}{2^{k}-1}}$.
\STATE Compute $r \leftarrow U \cdot Z$.
\STATE Compute index via inverse CDF: $i \leftarrow \lfloor \smash{\frac{\log_2(1 + r(2^k-1))}{k} - 1 \rfloor}$.
\STATE Clip $i$ to $[0, N_{\text{current}}-1]$.
\STATE \textbf{return} transition $\mathcal{D}[j][i]$.

\end{algorithmic}
\label{alg:recency-sampling-parallel}
\end{algorithm}
\vspace{-0.15in}

\section{Experimental Details}
\label{app:experimental_details}

\subsection{Figures}
\paragraph{\cref{fig:frontpage}.}
 We report the results in \emph{\textbf{Low \Volumename, High \Densityname}} setting on HumanoidBench~\citep{sferrazza2024humanoidbench}, with FastTD3~\citep{FastTD3} evaluated across all 29 Hands tasks and BRC~\citep{nauman2025bigger} across all 20 available tasks. Both algorithms operate in the low \text{\volumename} regime --- FastTD3~\citep{FastTD3} with a UTD ratio of $1/64$ and BRC~\citep{nauman2025bigger} with $1/10$ --- where recency bias is expected to yield the greatest benefit (Section~\ref{sec:sampling_scale}). Wall-clock time is measured as the total running time of the algorithm and is aggregated across both settings, covering 49 tasks in total.

\paragraph{\cref{fig:sampling-distributions}.}
We show expected amount of replays we sample from each index in between each environment step. Index recency is normalized index in replay buffer with 0 corresponding to the oldest sample and 1 corresponding to the newest sample in the buffer. By default all figures correspond to expected recency of training with UTD 2 and Batch Size 128.
On left and middle plot mean of Uniform sampling distribution, $\mu_\text{U}$ has index recency $\rho_t(i)=0.5$ and mean of Truncated Geometric distribution $\mu_\text{TG}$ has index recency $\rho_t(i)=0.857$. On middle plot higher Uniform \volumename{} corresponds to training with UTD 16 and Batch Size 128.
On right plot we provide numeric values of Entropy for Truncated Geometric distribution of size $1,000,000$ and Uniform FIFO distribution of size $288,000$. They both have the same mean $\mu_\text{TG}=\mu_\text{UF} \approx 0.85$.

\paragraph{\cref{fig:utd_entropy}.}
We report the results in varying \emph{\textbf{\text{\Volumename}}} comparing \emph{\textbf{High \text{\Densityname}}} Truncated Geometric replay and \emph{\textbf{Low \text{\Densityname}}} Uniform replay on HumanoidBench Nohands~\citep{sferrazza2024humanoidbench} using SimbaV2~\citep{SimbaV2} with 95\% stratified bootstrap CI computed via RLiable~\citep{RLiable}.  All experiments were run on $1,000,000$ steps with hyperparameters described in~\cref{tab:algorithm-hparams} in column 'SimbaV2'.

\paragraph{\cref{fig:timing_frontier}.} Results are reported in \emph{\textbf{Low \text{\Volumename}}} comparing \emph{\textbf{High \text{\Densityname}}} Truncated Geometric, ERE, and Uniform FIFO replays with \emph{\textbf{Low \text{\Densityname}}} Uniform replay on HumanoidBench~\citep{sferrazza2024humanoidbench} using BRC~\citep{nauman2025bigger}. All results have 95\% stratified bootstrap confidence intervals within $\pm 1.4\%$, computed with RLiable~\citep{RLiable}. The entropy-recency frontier shown in the right panel is computed for a buffer of capacity $1$M. Visualizations of the corresponding sampling distributions are provided in Figure~\ref{fig:distributions}. We discuss Uniform FIFO as a recency-biased strategy in Appendix~\ref{app:fifo}.

\paragraph{\cref{fig:latency}.} Measurements are conducted in \emph{\textbf{Low \text{\Volumename}}} comparing \emph{\textbf{High \text{\Densityname}}} Truncated Geometric replay with \textbf{Low \emph{\text{\Densityname}}} Uniform and PER replay on HumanoidBench using BRC across 20 tasks. PER maintains a separate segment tree per task, implemented in NumPy~\citep{harris2020array}, to avoid Q-value scale mismatch across tasks. Latency is measured using \texttt{time.time} in Python; each operation is timed over 1000 consecutive steps and averaged to obtain a reliable estimate. All measurements are taken approximately at the midpoint of training.

\paragraph{\cref{fig:learning_curve}.}
Results on both plots are reported in \emph{\textbf{Low \text{\Volumename}}} setting on HumanoidBench~\citep{sferrazza2024humanoidbench} comparing \emph{\textbf{High \text{\Densityname}}} Truncated Geometric and ERE replays with \emph{\textbf{Low \text{\Densityname}}} Uniform and PER replays. Large Scale Parallel Simulation uses FastTD3~\citep{FastTD3} algorithm. Multi-Task Learning uses BRC~\citep{nauman2025bigger}. Shaded regions are $95\%$ stratified bootstrap CI computed via RLiable~\citep{RLiable}. 
On Large Scale Parallel Simulation we report total number of environment steps on x-axis. We tuned Episode Length and Decay Constant $\eta$ in ERE buffer to achieve better results.

\paragraph{\cref{fig:ablations}.}
Results on $\alpha$ and actor/critic ablations are reported in \emph{\textbf{Low \text{\Volumename}}} setting on HumanoidBench~\citep{sferrazza2024humanoidbench}, results on distribution ablations are reported on DMC Dogs~\citep{DMC}. All ablations use BRC~\citep{nauman2025bigger} algorithm with $95\%$ stratified bootstrap CI computed via RLiable~\citep{RLiable}. Details of sampling distributions from right ablation are discussed in~\cref{sec:ablation_details}.

\subsection{Hyperparameters}
\label{app:hyperparams}

Table~\ref{tab:algorithm-hparams} reports the full set of hyperparameters used across all experiments. For each algorithm, we distinguish between default hyperparameters (shown in \textbf{bold}), which are taken directly from the original implementations, and tuned hyperparameters (shown in \textit{italic}), which were adjusted to accommodate the different replay sampling strategies.

\begin{table}[H]
\centering
\resizebox{\linewidth}{!}{
\begin{tabular}{l c c c c}
\toprule
\textbf{Hyperparameter} 
& \textbf{FastTD3} 
& \textbf{BRC} 
& \textbf{SimbaV2}
& \textbf{DQN} \\
\midrule
\multicolumn{5}{l}{\textit{Environment and Execution}} \\
Environment suite & HumanoidBench/Isaac Lab & HumanoidBench/DMC & \textbf{HumanoidBench Nohands}/DMC/MYOSuite & \textbf{Atari} \\
Parallel environments & $\boldsymbol{128}$ \textbf{(HumanoidBench)}/$\boldsymbol{2048}$ \textbf{(Isaac Lab)} & $\boldsymbol{20}$ \textbf{(HB)}, $\boldsymbol{3}$ \textbf{(DMC/Humanoids)}, $\boldsymbol{4}$ \textbf{(DMC Dogs)} & $-$ & $\boldsymbol{1}$ \\
Evaluation environments & $\boldsymbol{128}$ \textbf{(HumanoidBench)}/$\boldsymbol{2048}$ \textbf{(Isaac Lab)} & $-$ & $-$ & $-$ \\
Total environment steps & $\boldsymbol{100{,}000}$ & $\boldsymbol{1{,}000{,}000}$ & $\boldsymbol{1{,}000{,}000}$ & $\boldsymbol{10{,}000{,}000}$ \\
Number of seeds & $\boldsymbol{5}$ & $\boldsymbol{3\text{--}6}$ & $\boldsymbol{5}$ & $\boldsymbol{5}$ \\
TD-n & $\boldsymbol{1}$ \textbf{(HumanoidBench)}/$\boldsymbol{1\text{--}8}$ \textbf{(Isaac Lab)} & $\boldsymbol{1}$ & $\boldsymbol{1}$ & $\boldsymbol{1}$ \\
\midrule
\multicolumn{5}{l}{\textit{Optimization}} \\
Update-To-Data ratio & $\boldsymbol{1/64}$ \textbf{(HumanoidBench)}/$\boldsymbol{1/128}$ \textbf{(Isaac Lab)} & $\boldsymbol{1/10}$ & $\boldsymbol{2}, 1, 1/2, 1/4, 1/8, 1/16$ & $\boldsymbol{1/4}$ \\
Actor learning rate & $\boldsymbol{3\times10^{-4}}$ & $\boldsymbol{3\times10^{-4}}$ & $\boldsymbol{1\times10^{-4}}$ & $-$ \\
Critic learning rate & $\boldsymbol{3\times10^{-4}}$ & $\boldsymbol{3\times10^{-4}}$ & $\boldsymbol{1\times10^{-4}}$ & $\boldsymbol{1\times10^{-4}}$ \\
Learning rate schedule & \textbf{Cosine annealing} $\boldsymbol{(3\times10^{-4} \rightarrow 3\times10^{-5})}$ & \textbf{Constant} & \textbf{Linear annealing} $\boldsymbol{(1\times10^{-4} \rightarrow 5\times10^{-5})}$ & \textbf{Constant} \\
Batch size & $\boldsymbol{32768}$ & $\boldsymbol{1024}$ & $\boldsymbol{256}$ & $\boldsymbol{32}$ \\
Discount factor $\gamma$ & $\boldsymbol{0.99}$ & $\boldsymbol{0.99}$ & $\boldsymbol{0.95}$ \textbf{(HumanoidBench/DMC)}, $\boldsymbol{0.995}$ \textbf{(MYOSuite)} & $\boldsymbol{0.99}$ \\
Target update coefficient $\tau$ & $\boldsymbol{0.1}$ & $\boldsymbol{0.005}$ & $\boldsymbol{0.005}$ & $\boldsymbol{1.0}$ \textbf{(every 1000 steps)} \\
Policy update frequency & $\boldsymbol{2}$ & $\boldsymbol{2}$ & $\boldsymbol{2}$ & $-$ \\
Learning starts & $\boldsymbol{10}$ \textbf{steps} & $\boldsymbol{5000}$ \textbf{steps} & $\boldsymbol{5000}$ & $\boldsymbol{80{,}000}$ \textbf{steps} \\
Weight decay & $\boldsymbol{0.1}$ & $-$ & $-$ & $-$ \\
\midrule
\multicolumn{5}{l}{\textit{Replay Buffer}} \\
Replay buffer capacity & $\boldsymbol{50{,}000}$ \textbf{(HumanoidBench)} / $\boldsymbol{5 \times 1024 - 10 \times 1024}$ \textbf{(Isaac Lab)} & $100{,}000$, $\boldsymbol{1{,}000{,}000}$ & $\boldsymbol{1{,}000{,}000}$ & $\boldsymbol{1{,}000{,}000}$ \\
Sampling strategies & \textbf{Uniform} / PER / ERE / TG & \textbf{Uniform} / PER / ERE / TG & \textbf{Uniform} / PER / ERE / TG & \textbf{Uniform} / TG \\
\midrule
\multicolumn{5}{l}{\textit{Method-Specific Hyperparameters}} \\
Truncated Geometric $\alpha$ & $\boldsymbol{10.0}$ & $\boldsymbol{\mathit{10.0}}$ & $\boldsymbol{10.0}$ & $\boldsymbol{10.0}$ \\
PER exponent $\alpha$ & $\boldsymbol{0.6}$ & $\mathit{0.4}$ & $\boldsymbol{0.6}$ & $-$ \\
PER importance-sampling $\beta$ & $\boldsymbol{0.4 \rightarrow 1.0}$ & $\boldsymbol{0.4 \rightarrow 1.0}$ & $\boldsymbol{0.4 \rightarrow 1.0}$ & $-$ \\
ERE decay constant $\eta$ & $\boldsymbol{0.996}$ & $\boldsymbol{\mathit{0.996}}$ \textbf{(HB, DMC Dogs)}, $\boldsymbol{\mathit{0.994}}$ \textbf{(DMC Humanoids)} & $0.994$ & $-$ \\
ERE episode length & $\mathit{100}$ & $\boldsymbol{1000}$ & $\boldsymbol{1000}$ & $-$ \\
ERE mini-batch updates & $\mathit{200}$ & $\boldsymbol{2000}$ & $\boldsymbol{2000}$ & $-$ \\
ERE minimum buffer size $c_{\min}$ & $\boldsymbol{10 \times \text{batch size}}$ & $10 \times \text{batch size}$ & $5000$ & $-$ \\
\midrule
\multicolumn{5}{l}{\textit{Architecture}} \\
Actor hidden dimension & $\boldsymbol{512}$ & $\boldsymbol{256}$ & $\boldsymbol{128}$ & $-$ \\
Critic hidden dimension & $\boldsymbol{1024}$ & $\boldsymbol{4096}$ & $\boldsymbol{512}$ & $\boldsymbol{512}$ \textbf{(Nature CNN)} \\
Critic blocks & $\boldsymbol{2}$ & $\boldsymbol{2}$ & $\boldsymbol{2}$ & $-$ \\
Actor blocks & $\boldsymbol{1}$ & $\boldsymbol{1}$ & $\boldsymbol{1}$ & $-$ \\
Distributional atoms & $\boldsymbol{101}$ & $\boldsymbol{101}$ & $\boldsymbol{101}$ & $-$ \\
Value support $(v_{\min}, v_{\max})$ & $\boldsymbol{(-250, 250)}$ \textbf{(HumanoidBench)} / $\boldsymbol{(-50,50),(-10,10)}$ & $\boldsymbol{(-10, 10)}$ & $\boldsymbol{(-5,5)}$ & $-$ \\
Task embedding dimension & $-$ & $\boldsymbol{32}$ & $-$ & $-$ \\
\midrule
\multicolumn{5}{l}{\textit{Stability and Systems}} \\
Observation normalization & \textbf{Enabled} & \textbf{Disabled} & \textbf{Disabled} & \textbf{Disabled} \\
Reward normalization & \textbf{Disabled} & \textbf{Enabled} & \textbf{Enabled} & \textbf{Disabled} \\
Clipped Double Q & \textbf{Enabled} & \textbf{Disabled} & \textbf{Enabled} & \textbf{Disabled} \\
AMP & \textbf{Enabled (bf16)} & $-$ & $-$ & $-$ \\
\texttt{torch.compile} & \textbf{Enabled} (Excluding ERE) & $-$ & $-$ & $-$ \\
Gradient clipping & \textbf{Disabled} & \textbf{Disabled} & \textbf{Disabled} & \textbf{Disabled} \\
\bottomrule
\end{tabular}
}
\caption{
Algorithm-specific training and replay hyperparameters. FastTD3 values are taken directly from the default configuration (\texttt{BaseArgs}). BRC values are taken from BRC flag defaults. SimbaV2 is default configuration (\texttt{configs}). DQN values are the CleanRL defaults (\texttt{dqn\_atari\_jax.py}) with no changes; for DQN, critic rows refer to the Q-network, and exploration uses $\epsilon$-greedy with $\epsilon$ annealed linearly from $1.0$ to $0.01$ over the first $10\%$ of training. \textbf{Bold} denotes default hyperparameters, \textit{italic} denotes tuned hyperparameters.
}
\label{tab:algorithm-hparams}
\end{table}

\newpage
\subsection{Sampling Distributions}
\label{app:sampling_distributions}
\begin{figure}[H]
    \centering
    \vspace{-0.15in}
    \includegraphics[width=1.0\linewidth]{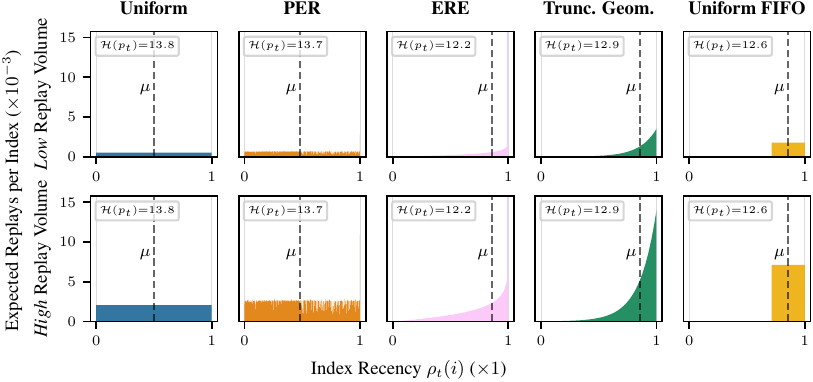}
    \vspace{-0.2in}
    \caption{\emph{\textbf{Sampling distributions and expected replay counts across methods.}}
Each column shows the sampling distribution of a replay strategy visualized as expected number of replays per buffer index, in the \emph{low} (top) and \emph{high} (bottom) replay volume regimes. The dashed vertical line denotes the \text{\densityname{}} $\mu$ for each strategy, and $\mathcal{H}(p_t)$ reports the sampling entropy. ERE, Truncated Geometric, and Uniform FIFO are matched to the same expected recency $\mu$, yet differ noticeably in their sampling entropy. Truncated Geometric achieves the highest entropy at this $\mu$. We discuss the interpretation of entropy differences in Appendix~\ref{app:entropy_scale}.}
    \label{fig:distributions}
    \vspace{-0.5cm}
\end{figure}

We visualize example sampling distributions for all used replay strategies in Figure~\ref{fig:distributions}. The Truncated Geometric, ERE, and Uniform FIFO distributions are matched to the same \text{\densityname{}} $\mu$ as used in Section~\ref{sec:sampling_entropy}.

\newpage
\subsection{Details of Distributions From Ablations} \label{sec:ablation_details}
In distribution ablation from~\cref{fig:ablations} we train BRC~\citep{nauman2025bigger} with different sampling distributions:
\begin{itemize}
\item \emph{\textbf{High \Densityname}} Linear.
\item \emph{\textbf{Low \Densityname}} Reverse Linear.
\item \emph{\textbf{Low \Densityname}} Gaussian.
\item \emph{\textbf{Low \Densityname}} Reverse Truncated Geometric.

\end{itemize}

\begin{figure}[H]
    \centering
    \includegraphics{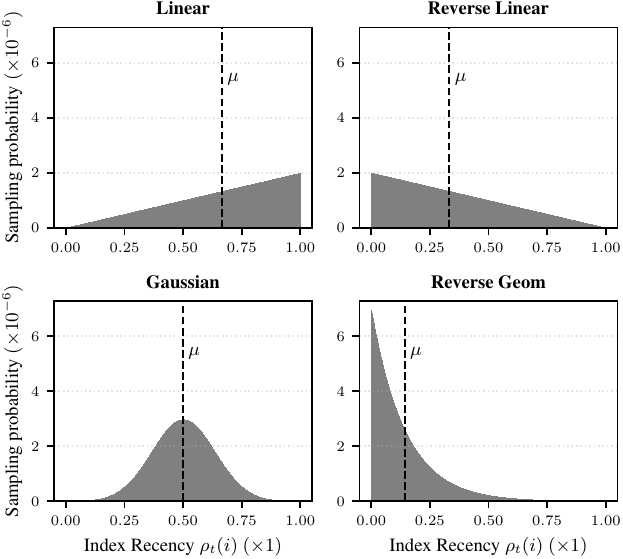}
    \caption{\textbf{Sampling distributions from Figure~\ref{fig:ablations}.} All four distributions are normalized so that the highest-probability transition is $2^{10}$ times more likely to be sampled than the lowest-probability transition.}
    \label{fig:diff_dist_sampling_curve}
    
\end{figure}

\label{app:distributions}

\subsection{Computational Resources}
\label{app:resources}
All experiments were conducted on NVIDIA A100 GPUs. Single-task and large-scale parallel runs required approximately 3 hours to complete, while multi-task runs required approximately 40 hours.

\subsection{Reproducibility}

All environment lists, hyperparameter configurations,  and training scripts are provided. Code is available at \url{https://github.com/M-Korniak/when-does-replay-matter}.

\section{Evaluation Protocols}
\label{app:training_details}

This section provides full details of the experimental setup used throughout the paper, including replay buffer configurations, algorithm-specific hyperparameters, normalization procedures, and evaluation protocols. All experiments are conducted using the original implementations of the corresponding algorithms, with minimal modifications required to integrate alternative replay sampling strategies.

\subsection{Policy Evaluation}

All policies are evaluated periodically during training without exploration noise. Each evaluation consists of multiple episodes per task, and results are averaged. All reported curves show the mean performance over random seeds, with $95\%$ confidence intervals computed using nonparametric bootstrapping via RLiable~\citep{RLiable}. Unless otherwise specified, all experiments are run with three to six random seeds (see \Cref{tab:algorithm-hparams}).

\subsection{Benchmarks}
\label{app:task_lists}
In this section we describe benchmarks considered in our studies:
\subsubsection{Multi-task benchmarks}
\label{app:multi-task-benchmarks}
All tasks are available in BRC~\citep{nauman2025bigger} repository \url{https://github.com/naumix/BiggerRegularizedCategorical}.
\begin{itemize}
  \item \textbf{BRC HumanoidBench Hands} - we use all 20 tasks of the HumanoidBench-Hard benchmark from~\citep{nauman2025bigger}, built on~\citep{sferrazza2024humanoidbench}. 
\begin{quote}
Tasks: \emph{walk, stand, run, stair, crawl, pole, slide, hurdle, maze, sit\_simple, sit\_hard, balance\_simple, balance\_hard, reach, spoon, window, insert\_small, insert\_normal, bookshelf\_simple, bookshelf\_hard}
\end{quote}

\item \textbf{BRC Meta-World} - we use benchmark of all 50 tasks from~\citep{metaworld}.
\begin{quote}
Tasks: \emph{ assembly-v2-goal-observable, basketball-v2-goal-observable, bin-picking-v2-goal-observable, box-close-v2-goal-observable, button-press-topdown-v2-goal-observable, button-press-topdown-wall-v2-goal-observable, button-press-v2-goal-observable, button-press-wall-v2-goal-observable, coffee-button-v2-goal-observable, coffee-pull-v2-goal-observable, coffee-push-v2-goal-observable, dial-turn-v2-goal-observable, disassemble-v2-goal-observable, door-close-v2-goal-observable, door-lock-v2-goal-observable, door-open-v2-goal-observable, door-unlock-v2-goal-observable, hand-insert-v2-goal-observable, drawer-close-v2-goal-observable, drawer-open-v2-goal-observable, faucet-open-v2-goal-observable, faucet-close-v2-goal-observable, hammer-v2-goal-observable, handle-press-side-v2-goal-observable, handle-press-v2-goal-observable, handle-pull-side-v2-goal-observable, handle-pull-v2-goal-observable, lever-pull-v2-goal-observable, pick-place-wall-v2-goal-observable, pick-out-of-hole-v2-goal-observable, pick-place-v2-goal-observable, plate-slide-v2-goal-observable, plate-slide-side-v2-goal-observable, plate-slide-back-v2-goal-observable, plate-slide-back-side-v2-goal-observable, peg-insert-side-v2-goal-observable, peg-unplug-side-v2-goal-observable, soccer-v2-goal-observable, stick-push-v2-goal-observable, stick-pull-v2-goal-observable, push-v2-goal-observable, push-wall-v2-goal-observable, push-back-v2-goal-observable, reach-v2-goal-observable, reach-wall-v2-goal-observable, shelf-place-v2-goal-observable, sweep-into-v2-goal-observable, sweep-v2-goal-observable, window-open-v2-goal-observable, window-close-v2-goal-observable}
\end{quote}
\item \textbf{BRC DMC Dogs} - we use benchmark of all 4 DMC Dogs tasks~\citep{DMC}.
\begin{quote}
Tasks: \emph{dog-stand, dog-walk, dog-trot, dog-run}
\end{quote}
\item \textbf{BRC DMC Humanoids} - we use benchmark of all 3 DMC Humanoid tasks~\citep{DMC}.
\begin{quote}
Tasks: \emph{humanoid-stand, humanoid-walk, humanoid-run}
\end{quote}

\end{itemize}
\subsubsection{Large Scale Parallel Learning benchmarks}
All tasks are available in FastTD3~\citep{FastTD3} repository \url{https://github.com/younggyoseo/FastTD3}
\begin{itemize}
\item \textbf{FastTD3 HumanoidBench Hands} - we use benchmark of 29 HumanoidBench Hands tasks from~\citep{sferrazza2024humanoidbench}
\begin{quote}
Tasks: \emph{balance\_hard, balance\_simple, basketball, bookshelf\_hard, bookshelf\_simple, cabinet, crawl, cube, door, hurdle, insert\_normal, insert\_small, maze, package, pole, powerlift, push, reach, room, run, sit\_hard, sit\_simple, slide, spoon, stair, stand, truck, walk, window}
\end{quote}
\item \textbf{FastTD3 Isaac Lab} - we use benchmark of 6 Isaac Lab~\citep{isaaclab} tasks
\begin{quote}
Tasks: \emph{Velocity-Flat-G1, Velocity-Flat-H1, Velocity-Rough-G1, Velocity-Rough-H1, Repose-Cube-Allegro-Direct, Repose-Cube-Shadow-Direct}
\end{quote}
\end{itemize}
\subsubsection{Single-task benchmarks}
All tasks are available for use in SimbaV2~\citep{SimbaV2} repository \url{https://github.com/DAVIAN-Robotics/SimbaV2}
\begin{itemize}
\item \textbf{SimbaV2 HumanoidBench No Hands} - we use benchmark of 13 HumanoidBench No Hands tasks from~\citep{sferrazza2024humanoidbench}.
\begin{quote}
Tasks: \emph{walk, stand, run, hurdle, crawl, maze, sit\_simple, sit\_hard, balance\_simple, balance\_hard, stair, slide, pole}
\end{quote}
\item \textbf{SimbaV2 MyoSuite} - we use benchmark of 10 MyoSuite tasks from~\citep{caggiano2022myosuite}.
\begin{quote}
Tasks: \emph{reach, reach-hard, pose, pose-hard, obj-hold, obj-hold-hard, key-turn, key-turn-hard, pen-twirl, pen-twirl-hard}
\end{quote}
\item \textbf{SimbaV2 DMC} - we use 7 tasks from combined DMC Dogs and DMC Humanoids from~\citep{DMC}.
\begin{quote}
Tasks: \emph{humanoid-stand, humanoid-walk, humanoid-run, dog-stand, dog-walk, dog-trot, dog-run}
\end{quote}
\end{itemize}

\subsubsection{Discrete-action benchmarks}
All games are available through the Arcade Learning Environment~\citep{bellemare2013arcade}; we use the DQN implementation from the CleanRL~\citep{huang2022cleanrl} repository \url{https://github.com/vwxyzjn/cleanrl}
\begin{itemize}
\item \textbf{CleanRL DQN Atari} - we use 56 Atari games (the standard Atari-57 set without \emph{Surround}, which is unavailable in the \texttt{NoFrameskip-v4} environments).
\begin{quote}
Tasks: \emph{Alien, Amidar, Assault, Asterix, Asteroids, Atlantis, BankHeist, BattleZone, BeamRider, Berzerk, Bowling, Boxing, Breakout, Centipede, ChopperCommand, CrazyClimber, Defender, DemonAttack, DoubleDunk, Enduro, FishingDerby, Freeway, Frostbite, Gopher, Gravitar, Hero, IceHockey, Jamesbond, Kangaroo, Krull, KungFuMaster, MontezumaRevenge, MsPacman, NameThisGame, Phoenix, Pitfall, Pong, PrivateEye, Qbert, Riverraid, RoadRunner, Robotank, Seaquest, Skiing, Solaris, SpaceInvaders, StarGunner, Tennis, TimePilot, Tutankham, UpNDown, Venture, VideoPinball, WizardOfWor, YarsRevenge, Zaxxon}
\end{quote}
\end{itemize}
\subsection{Normalization and Metrics}

Performance is evaluated using standard task-specific metrics, including normalized episodic returns and success rates where applicable. Normalized returns are computed as
\[
\text{Score} = \frac{R - R_{\text{random}}}{R_{\text{success}} - R_{\text{random}}},
\]
where $R_{\text{random}}$ and $R_{\text{success}}$ are task-specific constants provided by the benchmark.

Table~\ref{tab:normalization_constants} reports normalization constants and success thresholds for all tasks sourced from \citet{sferrazza2024humanoidbench}.

\begin{table}[h!]
\centering

\begin{tabular}{l r r}
\toprule
\textbf{Task} & \textbf{Random Score $R_{\text{random}}$} & \textbf{Success Threshold $R_{\text{success}}$} \\
\midrule
h1-balance\_hard-v0 & 9.044 & 800.0 \\
h1-balance\_simple-v0 & 9.391 & 800.0 \\
h1-crawl-v0 & 272.658 & 700.0 \\
h1-hurdle-v0 & 2.214 & 700.0 \\
h1-maze-v0 & 106.441 & 1200.0 \\
h1-pole-v0 & 20.090 & 700.0 \\
h1-reach-v0 & 260.302 & 12000.0 \\
h1-run-v0 & 2.020 & 700.0 \\
h1-sit\_hard-v0 & 2.448 & 750.0 \\
h1-sit\_simple-v0 & 9.393 & 750.0 \\
h1-slide-v0 & 3.191 & 700.0 \\
h1-stair-v0 & 3.112 & 700.0 \\
h1-stand-v0 & 10.545 & 800.0 \\
h1-walk-v0 & 2.377 & 700.0 \\
\midrule
h1hand-balance\_hard-v0 & 10.032 & 800.0 \\
h1hand-balance\_simple-v0 & 10.170 & 800.0 \\
h1hand-basketball-v0 & 8.979 & 1200.0 \\
h1hand-bookshelf\_hard-v0 & 14.848 & 2000.0 \\
h1hand-bookshelf\_simple-v0 & 16.777 & 2000.0 \\
h1hand-cabinet-v0 & 37.733 & 2500.0 \\
h1hand-crawl-v0 & 278.868 & 800.0 \\
h1hand-cube-v0 & 4.787 & 370.0 \\
h1hand-door-v0 & 2.771 & 600.0 \\
h1hand-hurdle-v0 & 2.371 & 700.0 \\
h1hand-insert\_normal-v0 & 1.673 & 350.0 \\
h1hand-insert\_small-v0 & 1.653 & 350.0 \\
h1hand-kitchen-v0 & 0.000 & 4.0 \\
h1hand-maze-v0 & 106.233 & 1200.0 \\
h1hand-package-v0 & -10040.932 & 1500.0 \\
h1hand-pole-v0 & 19.721 & 700.0 \\
h1hand-powerlift-v0 & 17.638 & 800.0 \\
h1hand-push-v0 & -526.800 & 700.0 \\
h1hand-reach-v0 & -50.024 & 12000.0 \\
h1hand-room-v0 & 3.018 & 400.0 \\
h1hand-run-v0 & 1.927 & 700.0 \\
h1hand-sit\_hard-v0 & 2.477 & 750.0 \\
h1hand-sit\_simple-v0 & 10.768 & 750.0 \\
h1hand-slide-v0 & 3.142 & 700.0 \\
h1hand-spoon-v0 & 4.661 & 650.0 \\
h1hand-stair-v0 & 3.161 & 700.0 \\
h1hand-stand-v0 & 11.973 & 800.0 \\
h1hand-truck-v0 & 562.419 & 3000.0 \\
h1hand-walk-v0 & 2.505 & 700.0 \\
h1hand-window-v0 & 2.713 & 650.0 \\
\bottomrule
\end{tabular}

\caption{Normalization constants ($R_{\text{random}}$) and success thresholds ($R_{\text{success}}$) for HumanoidBench tasks.}
\label{tab:normalization_constants}
\end{table}

\newpage
\section{Discussion}
\label{app:discussion}

\subsection{Normalized AUC for Sample Efficiency}
\label{app:auc}

To evaluate sample efficiency, we report the \emph{Normalized Area Under the Curve} (Normalized AUC). Rather than measuring performance only at the final training step, this metric captures learning progress throughout training and relates to the cumulative regret that the agent incurs during learning.

Formally, let $R_t$ denote the normalized return at evaluation step $t$. Given equally spaced evaluation intervals $\{t_1, \dots, t_K\}$, we compute

\[
\text{Normalized AUC} = \frac{1}{K} \sum_{k=1}^{K} R_{t_k}.
\]

For each run and task, we first compute the temporal mean of normalized returns across evaluation checkpoints. 
We then average these per-task Normalized AUC values across tasks and finally report the mean across runs. Confidence intervals are estimated using stratified bootstrap resampling over runs, following the RLiable protocol~\citep{RLiable}.

Normalized AUC provides a more faithful measure of sample efficiency than reporting only the final performance. Final-step metrics can obscure differences in convergence speed and learning stability, whereas Normalized AUC captures both how quickly and how consistently an agent improves during training.

This evaluation protocol follows prior work that emphasizes area-under-curve metrics for sample-efficient reinforcement learning~\citep{RLiable}.

\subsection{Uniform FIFO as a Recency-Biased Method}
\label{app:fifo}

A Uniform FIFO replay buffer can be interpreted as a recency-biased method. At each step, the oldest transition is discarded, which is equivalent to assigning it a sampling probability of zero. In this sense, FIFO implements recency bias through the discarding of oldest transitions rather than through non-uniform sampling probabilities. 

Interestingly, in our preliminary experiments, alternative discarding strategies — such as discarding transitions based on Q-values, estimated advantages, or novelty estimated via random network distillation — did not improve over standard FIFO. This suggests that FIFO may be a surprisingly strong discarding baseline, consistent with the broader literature~\citep{wang2019boostingsoftactorcriticemphasizing, wang2020striving} suggesting that recency — implicitly induced by FIFO through discarding — is an important factor in replay buffer design. We leave a systematic investigation of discarding strategies as an interesting direction for future work.

\subsection{Entropy Scale and Buffer Size Equivalence}
\label{app:entropy_scale}

Since sampling entropy $\mathcal{H}(p_t)$ is a logarithmic quantity, small numerical differences correspond to substantial changes in sampling diversity. For intuition, consider a uniform distribution over a buffer of size $N$, which has entropy $\ln N$. A change of $1$ nat in entropy is therefore equivalent to scaling the effective buffer size by a factor of $e \approx 2.71$. Consequently, differences in entropy of $0.3$ nats --- which may appear modest --- correspond to a factor of $e^{0.3} \approx 1.35$ change in effective buffer size, i.e., a $35\%$ increase or decrease in sampling diversity.

An alternative metric is perplexity, defined as $e^{\mathcal{H}(p_t)}$, which trades the logarithmic scale of entropy for a linear one, potentially making scalar values easier to interpret directly. However, perplexity and entropy are bijectively equivalent, and since entropy is the more standard metric in the literature, we report entropy throughout this work. We encourage readers to keep the logarithmic scale in mind when interpreting entropy differences reported in our figures.

\subsection{ERE Buffer Shortcomings}
\label{app:ere_shortcomings}
On the left panel of \cref{fig:timing_frontier}, the ERE buffer~\citep{wang2019boostingsoftactorcriticemphasizing} underperforms severely. We believe this is caused by the low entropy of the distribution from which ERE samples. To make its \text{\densityname} match that of the Truncated Geometric with $\alpha=10$, we tune the Decay Constant $\eta$ by decreasing it. In the ERE buffer, replays are sampled in episodes $k$ increasing cyclically from $1$ to the episode length $K$; in each episode, we sample uniformly from $c_k=\max(N\cdot \eta^{1000k/K},c_{\text{min}})$ most recent samples, where $N$ is the length of the buffer and $c_{\text{min}}$ is the minimal sampling size, usually set to $5000$ for a buffer size of $1{,}000{,}000$. By decreasing the base of the exponent, i.e., the decay constant $\eta$, we increase the number of episodes for which we sample from the minimal window $c_k=c_{\text{min}}$; this drastically reduces \text{\entropyname}, as visible for high \text{\densityname} $\mu$ in the right subplot of \cref{fig:timing_frontier}. For the experiments in \cref{fig:learning_curve}, we tune the ERE parameters for every algorithm and benchmark individually to achieve competitive performance comparable to other methods. In comparison, Truncated Geometric replay used a fixed hyperparameter of $\alpha=10$ across all experiments.

\subsection{Choosing $\alpha$ in Truncated Geometric Sampler}
\label{app:alpha_choice}

We choose $\alpha$ so that the expected recency lies near the elbow of the
recency--entropy frontier (Figure~\ref{fig:timing_frontier}, right). This point
is recent enough to emphasize recent data, but comes before entropy starts to
fall sharply. In all experiments, we use $\alpha = 10$ with no per-benchmark or
per-task tuning. For a full buffer, this gives an expected recency of
$\mu \approx 0.85$, which lies near the elbow. This choice of $\alpha$ performs consistently
well across low-replay-volume settings (Section~\ref{sec:experiments}), and the
gains also hold for $\alpha \in \{4, 8\}$ (Figure~\ref{fig:ablations}, left). We
view $\alpha = 10$ as a practical, benchmark-independent heuristic rather than a
proven optimum; identifying the optimal $\alpha$ is left to future work.

\newpage
\section{Extended Results and Learning Curves }
\label{app:results}
\begin{figure}[H]
    \centering
    \includegraphics[width=\linewidth]{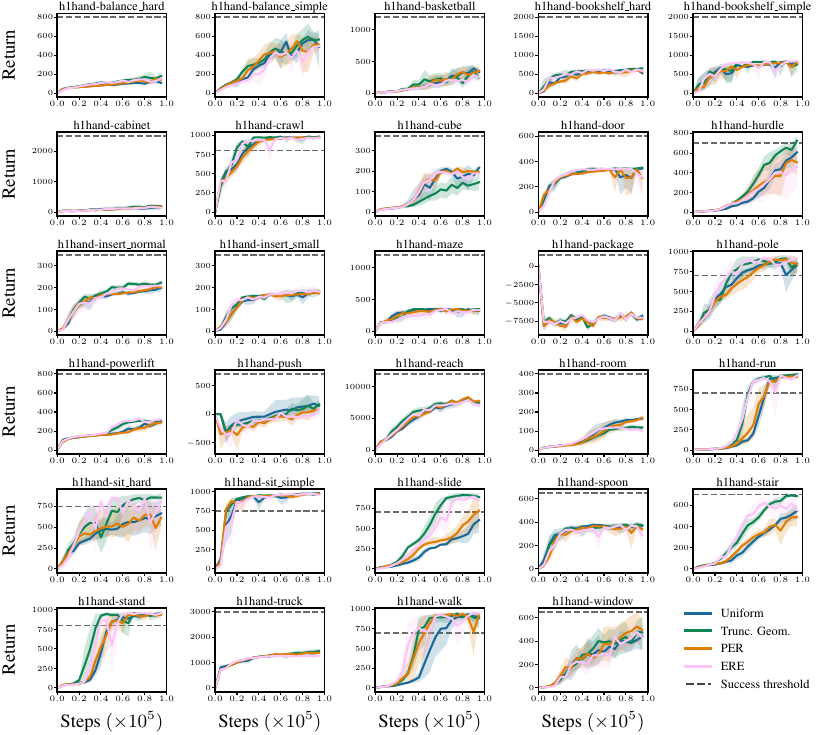}
    \caption{\emph{\textbf{(Low Replay Volume) Performance across all 29 tasks.}} We compare FastTD3~\citep{FastTD3} with different replay strategies: \emph{\textbf{Low \text{\densityname}}} - Uniform (blue), PER (orange) and \emph{\textbf{High \densityname}} - Truncated Geometric (green), and ERE (pink). Solid lines represent the mean return over 5 seeds, and shaded regions denote the $95\%$ bootstrap confidence intervals computed via \texttt{rliable} on HumanoidBench. The dashed grey line in each subplot indicates the task-specific success threshold. All x-axes represent environment steps ($\times 10^7$). On average, Truncated Geometric achieves a $14.1\%$ gain over Uniform in AUC.}
    \label{fig:fast-td3-per-task}
\end{figure}

\begin{figure}[H]
    \centering
    \includegraphics[width=\linewidth]{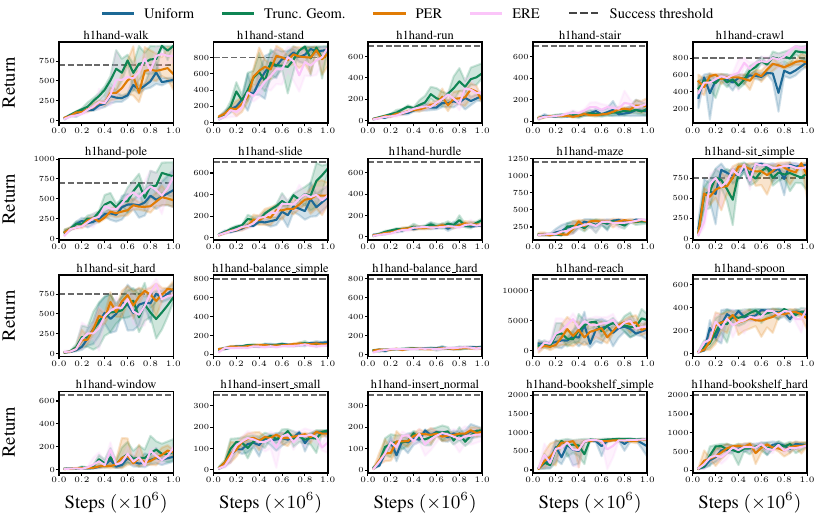}
    \caption{\emph{\textbf{(Low \text{\Volumename}}) Performance across all 20 BRC tasks.} We compare different replay strategies: \emph{\textbf{Low \text{\densityname}}} - Uniform (blue), PER (orange) and \emph{\textbf{High \densityname}} - Truncated Geometric (green), and ERE (pink). Solid lines represent the mean return over 5 seeds, and shaded regions denote the $95\%$ bootstrap confidence intervals computed via \texttt{rliable} on HumanoidBench. The dashed grey line in each subplot indicates the task-specific success threshold. All x-axes represent environment steps ($\times 10^6$).  On average, Truncated Geometric achieves a $12.9\%$ gain over Uniform in AUC and approximately $20\%$ improvement in final performance.}
    \label{fig:BRC-per-task}
\end{figure}

\begin{figure}[H]
    \centering
    \includegraphics[width=\linewidth]{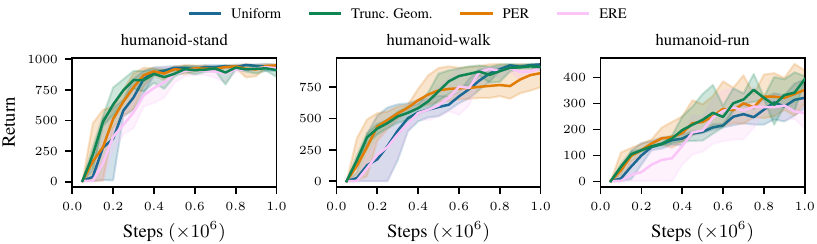}
    \caption{\emph{\textbf{(Moderate Replay Volume) Performance on DMC Humanoids tasks~\citep{DMC}}.} We compare different replay strategies: \emph{\textbf{Low \text{\densityname}}} - Uniform (blue), PER (orange) and \emph{\textbf{High \densityname}} - Truncated Geometric (green), and ERE (pink) across three humanoid tasks: \texttt{stand}, \texttt{walk}, and \texttt{run}. Solid lines show the mean return over all available seeds for each method, and shaded regions indicate $95\%$ bootstrap confidence intervals computed with \texttt{rliable}. All x-axes represent environment steps ($\times 10^6$), and y-axes show the denormalized task returns. On average, Truncated Geometric achieves a $10\%$ gain over Uniform in AUC.}
    \label{fig:dmc-humanoids-3-tasks}
\end{figure}

\begin{figure}[H]
    \centering
    \includegraphics[width=\linewidth]{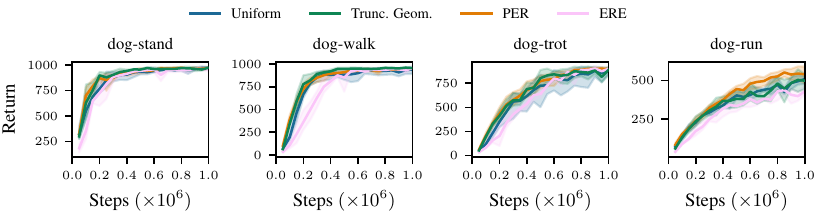}
    \caption{\looseness=-1\emph{\textbf{(Moderate Replay Volume) Performance on DMC Dog tasks.}} We compare different replay strategies on BRC~\citep{nauman2025bigger}: \emph{\textbf{Low \text{\densityname}}} - Uniform (blue), PER (orange) and \emph{\textbf{High \densityname}} - Truncated Geometric (green), and ERE (pink) across four dog locomotion tasks: \texttt{stand}, \texttt{walk}, \texttt{trot}, and \texttt{run}. Solid lines show the mean return over all available seeds for each method, and shaded regions indicate $95\%$ bootstrap confidence intervals computed with \texttt{rliable}. All x-axes represent environment steps ($\times 10^6$), and y-axes show the denormalized task returns. On average, Truncated Geometric achieves a $4.2\%$ gain over Uniform in AUC.}
    \label{fig:dmc-dogs-4-tasks}
\end{figure}

\begin{figure}[H]
    \centering
    \includegraphics[width=\linewidth]{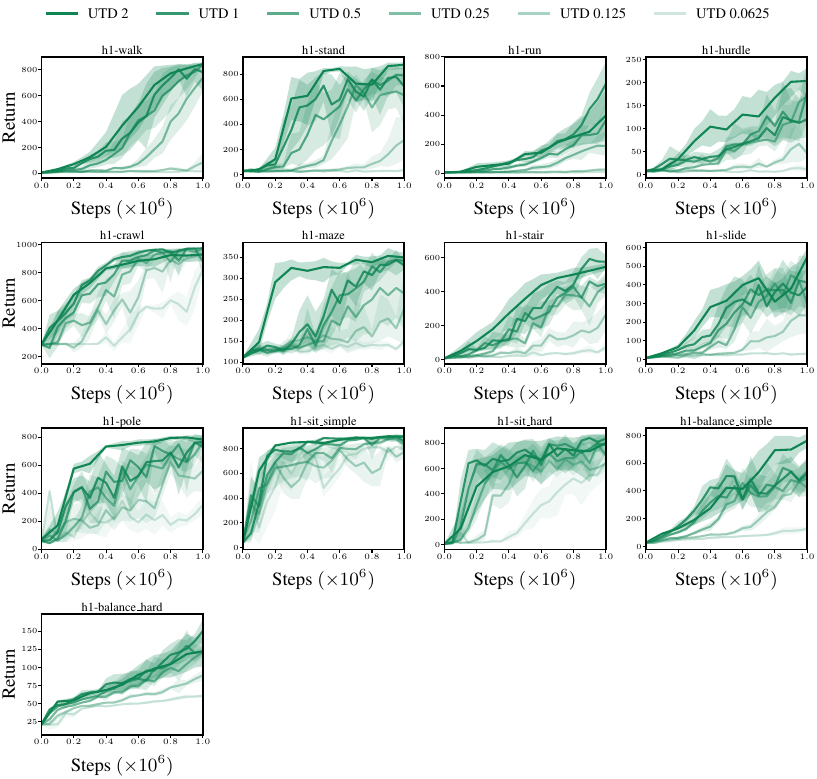}
    \caption{\emph{\textbf{(Changing \text{\Volumename}, High \text{\Densityname}) Performance on HumanoidBench Nohands tasks.}} We compare performance of SimbaV2 with Truncated Geometric replay for all UTDs reported in Figure~\ref{fig:utd_entropy}. Solid lines show the mean return over all available seeds for each method, and shaded regions indicate $95\%$ bootstrap confidence intervals computed with \texttt{rliable}. All x-axes represent environment steps ($\times 10^6$), and y-axes show the denormalized task returns.}
    \label{fig:hbench-truncgeom-utd-pertask}
\end{figure}

\begin{figure}[H]
    \centering
    \includegraphics[width=\linewidth]{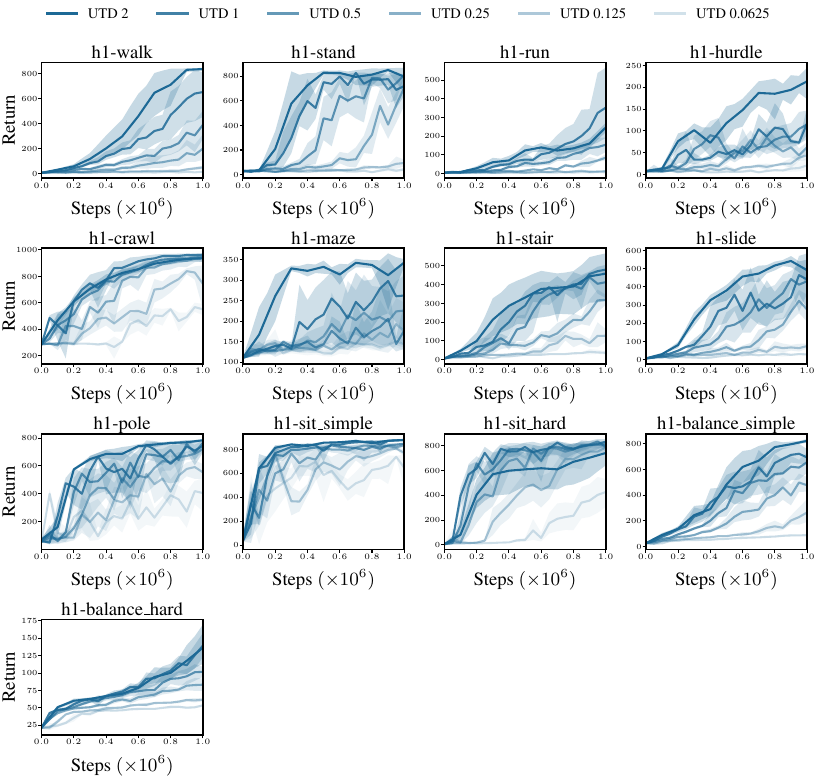}
    \caption{\emph{\textbf{(Changing \text{\Volumename}, Low \text{\Densityname}) Performance of SimbaV2 with Uniform replay on HumanoidBench Nohands tasks.}} We report performance for all UTDs from Figure~\ref{fig:utd_entropy}. Solid lines show the mean return over all available seeds for each method, and shaded regions indicate $95\%$ bootstrap confidence intervals computed with \texttt{rliable}. All x-axes represent environment steps ($\times 10^6$), and y-axes show the denormalized task returns.}
    \label{fig:hbench-uniform-utd-pertask}
\end{figure}

\begin{figure}[H]
    \centering
    \includegraphics[width=\linewidth]{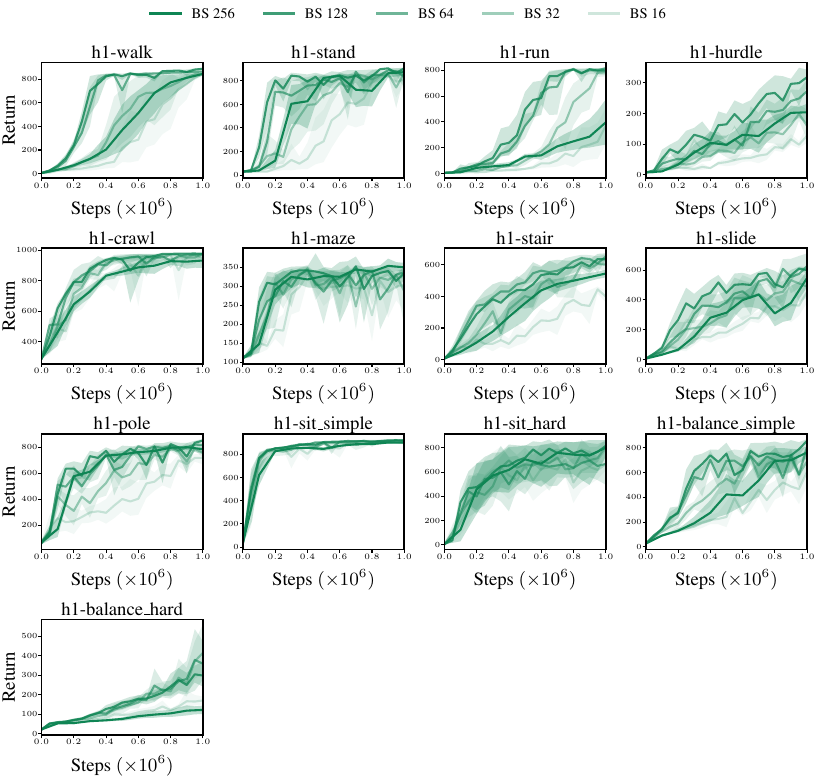}
    \caption{\emph{\textbf{(Changing \text{\Volumename}, High \text{\Densityname}) Performance on HumanoidBench Nohands tasks.}} We compare performance of SimbaV2 with Truncated Geometric replay for all batch sizes reported in Figure~\ref{fig:utd_entropy}. Solid lines show the mean return over all available seeds for each method, and shaded regions indicate $95\%$ bootstrap confidence intervals computed with \texttt{rliable}. All x-axes represent environment steps ($\times 10^6$), and y-axes show the denormalized task returns.}
    \label{fig:hbench-truncgeom-bs-pertask}
\end{figure}

\begin{figure}[H]
    \centering
    \includegraphics[width=\linewidth]{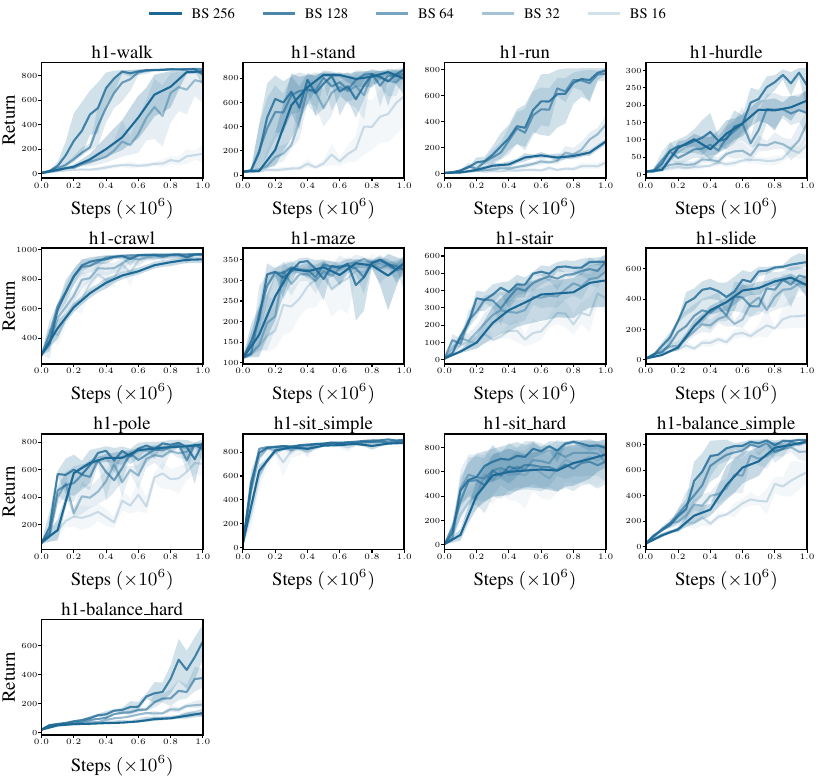}
    \caption{\emph{\textbf{(Changing \text{\Volumename}, Low \text{\Densityname}) Performance of SimbaV2 with Uniform replay on HumanoidBench Nohands tasks.}} We report performance for all batch sizes from Figure~\ref{fig:utd_entropy}. Solid lines show the mean return over all available seeds for each method, and shaded regions indicate $95\%$ bootstrap confidence intervals computed with \texttt{rliable}. All x-axes represent environment steps ($\times 10^6$), and y-axes show the denormalized task returns.}
    \label{fig:hbench-uniform-bs-pertask}
\end{figure}

We provide extended results and learning curves for the experiments presented in Sections~\ref{sec:analysis} and~\ref{sec:experiments}. In \Cref{fig:fast-td3-per-task} we present the performance of all sampling baselines when using FastTD3. In \Cref{fig:BRC-per-task,fig:dmc-humanoids-3-tasks,fig:dmc-dogs-4-tasks} we illustrate the learning curves for our experiments using BRC on HumanoidBench and DeepMind Control Suite. In Figures~\ref{fig:hbench-truncgeom-utd-pertask} to \ref{fig:hbench-per-bs-pertask} we illustrate the learning curves for experiment from \Cref{fig:utd_entropy}. In sections \ref{app:brc-Meta-World}, \ref{app:fasttd3-isaac} and \ref{app:atari} we report extra experiments mentioned in \Cref{sec:experiments}. In \Cref{app:constant-replay-simba} we report gains from SimbaV2 with Truncated Geometric replay compared to uniform replay. In \Cref{app:noisy_simba} we investigate behavior of SimbaV2 in Low Replay Volume when actions are perturbed randomly.

\subsection{Prioritized Experience Replay UTD and Batch Size ablation}
\label{app:per_utd_bs}
\begin{figure}[H]
    \centering
    \includegraphics[width=\linewidth]{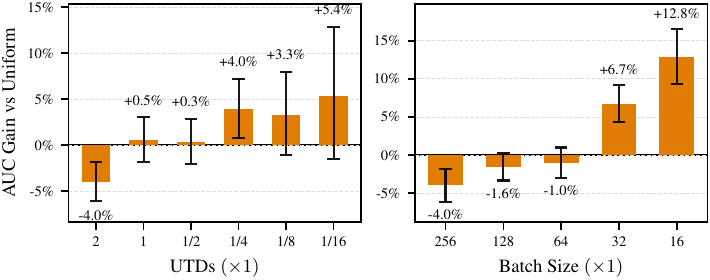}
    \caption{\emph{\textbf{SimbaV2 on HumanoidBench (13 tasks).}} Improvement of Prioritized Experience Replay over uniform replay as \textbf{\volumename} is varied through UTD 
    \emph{(left)} and batch size \emph{(right)}. We report a gain as a ratio of normalized AUCs from normalized returns over $5$ seeds for each method.
    Error bars show $95\%$ stratified bootstrap CIs computed with
    \texttt{rliable}.}
    \label{fig:auc_per}
\end{figure}

\newpage

\subsection{FastTD3}
\label{app:fasttd3_results}

\newpage
\subsection{BRC HumanoidBench}
\label{app:brc-hbench}

\subsection{BRC DeepMind Control Suite Humanoids}

\subsection{BRC DeepMind Control Suite  Dogs}
\subsection{SimbaV2 at default UTD learning curve}
\begin{figure}[H]
    \centering
    \includegraphics[width=\linewidth]{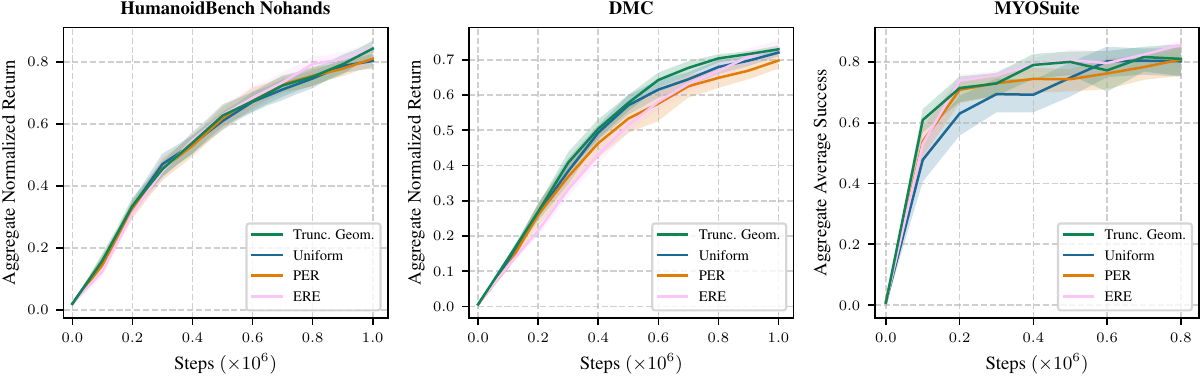}
    \caption{\emph{\textbf{(High Replay Volume) Performance on HumanoidBench Nohands tasks, DMC, and MYOSuite.}} We compare different replay strategies on SimbaV2~\citep{SimbaV2}: \emph{\textbf{Low \text{\densityname}}} - Uniform (blue), PER (orange) and \emph{\textbf{High \densityname}} - Truncated Geometric (green), and ERE (pink). We use UTD=2, standard UTD value taken from original paper on benchmarks HumanoidBench Nohands, DMC. Solid lines show the mean return over all available seeds for each method, and shaded regions indicate $95\%$ bootstrap confidence intervals computed with \texttt{rliable}. All x-axes represent environment steps ($\times 10^6$), and y-axes show the task returns. Truncated Geometric remains effectively neutral when replay volume is high.}
    \label{fig:simba-learningcurve}
\end{figure}

\newpage
\subsection{SimbaV2 UTD}

\newpage

\newpage
\begin{figure}[H]
    \centering
    \includegraphics[width=\linewidth]{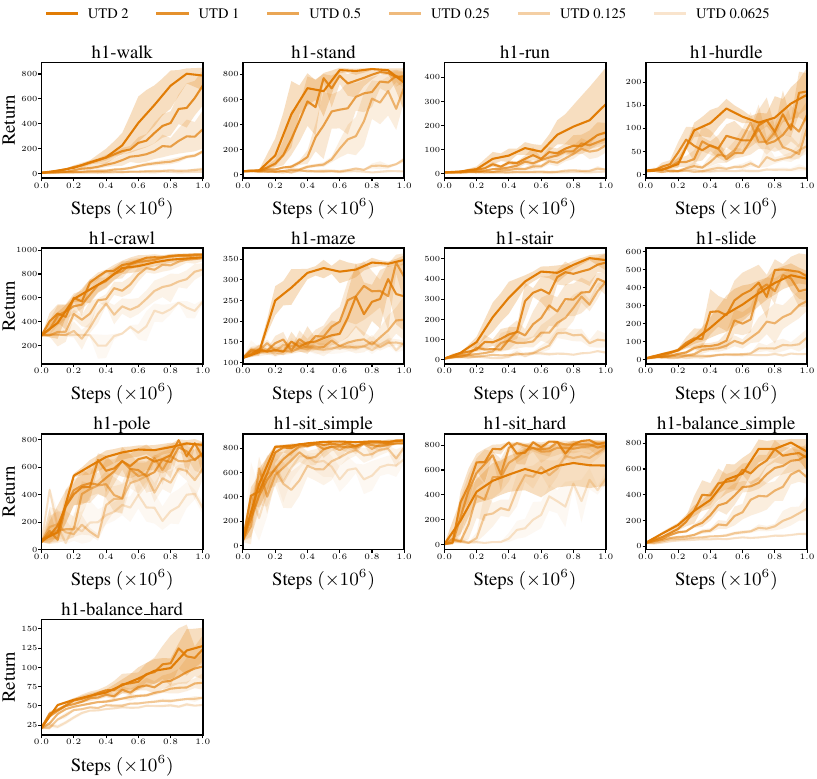}
    \caption{\emph{\textbf{(Changing \text{\Volumename}, Low \text{\Densityname}) Performance of SimbaV2 with Prioritized replay on HumanoidBench Nohands tasks.}} We report performance for all UTDs from Figure~\ref{fig:auc_per}. Solid lines show the mean return over all available seeds for each method, and shaded regions indicate $95\%$ bootstrap confidence intervals computed with \texttt{rliable}. All x-axes represent environment steps ($\times 10^6$), and y-axes show the denormalized task returns.}
    \label{fig:hbench-per-utd-pertask}
\end{figure}

\newpage
\subsection{SimbaV2 batch size}

\newpage

\newpage
\begin{figure}[H]
    \centering
    \includegraphics[width=\linewidth]{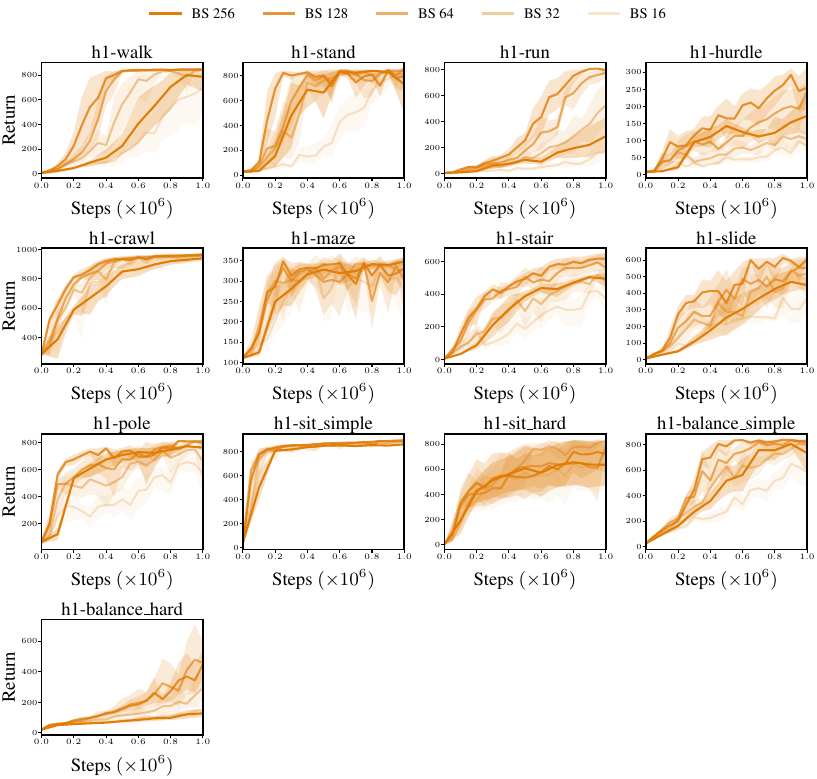}
    \caption{\emph{\textbf{(Changing \text{\Volumename}, Low \text{\Densityname}) Performance of SimbaV2 with Prioritized replay on HumanoidBench Nohands tasks.}} We report performance for all batch sizes from Figure~\ref{fig:auc_per}. Solid lines show the mean return over all available seeds for each method, and shaded regions indicate $95\%$ bootstrap confidence intervals computed with \texttt{rliable}. All x-axes represent environment steps ($\times 10^6$), and y-axes show the denormalized task returns.}
    \label{fig:hbench-per-bs-pertask}
\end{figure}

\newpage
\subsection{BRC Meta-World}
\label{app:brc-Meta-World}

\begin{figure}[H]
    \centering
    \includegraphics{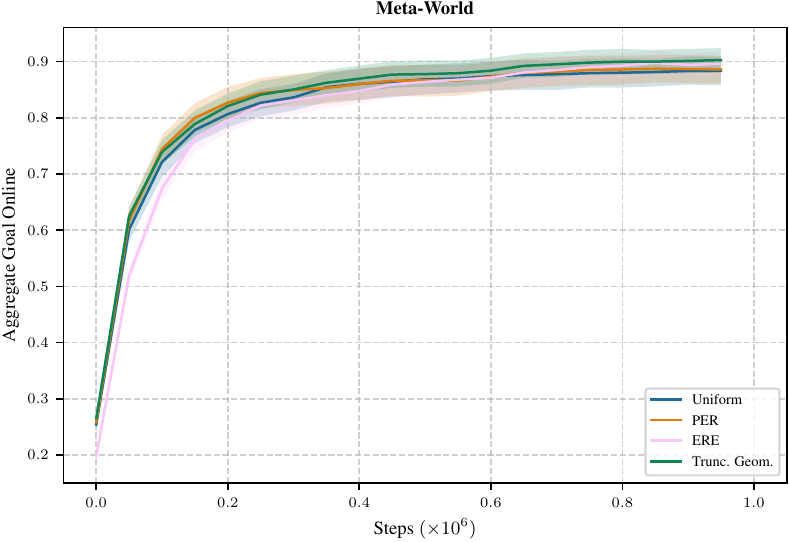}
    \caption{\emph{\textbf{(Low \text{\Volumename}) Mean performance of BRC~\citep{nauman2025bigger} on Meta-World}} with \emph{\textbf{Low \text{\densityname}}} - Uniform (blue), PER (orange) and \emph{\textbf{High \densityname}} - Truncated Geometric (green), and ERE (pink) replay strategies. Solid lines show the mean goal online over all available seeds for each method, and shaded regions indicate $95\%$ bootstrap confidence intervals computed with \texttt{rliable}. Truncated Geometric achieves a $2$ percentage point improvement over Uniform ($90\%$ vs.\ $88\%$ aggregate success rate), which is noticeable given that Meta-World is largely saturated at this performance level, leaving limited room for improvement.}
    \label{fig:Meta-World-multitask}
\end{figure}
\newpage
\begin{figure}[H]
    \centering
    \includegraphics[height=\textheight,width=\linewidth, keepaspectratio]{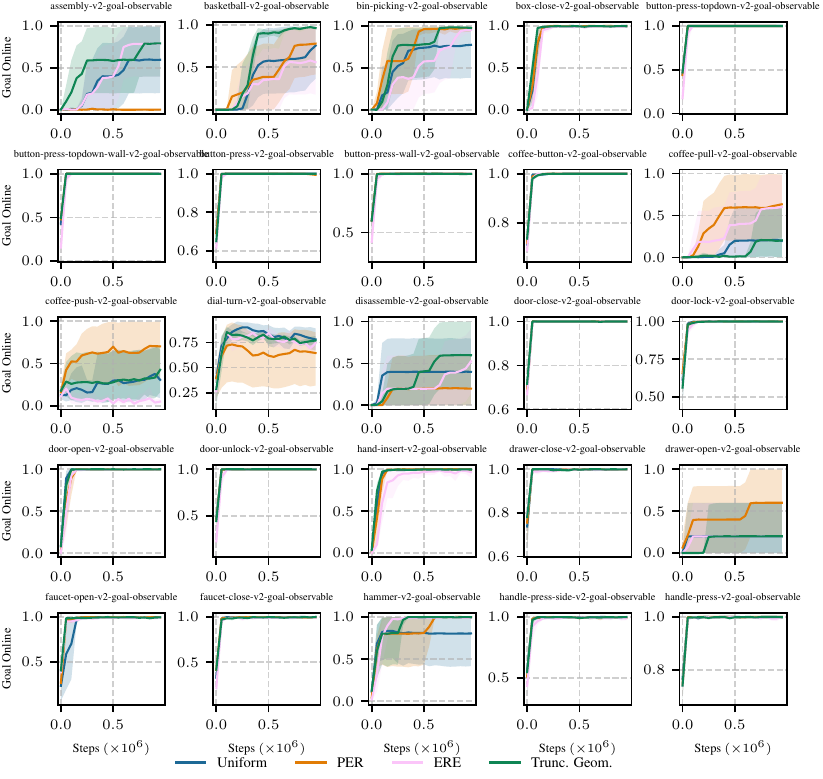}
    \caption{\emph{\textbf{(Low \text{\Volumename}) Mean performance of BRC~\citep{nauman2025bigger} with various replay strategies on Meta-World.}} 
    We use \emph{\textbf{Low \text{\densityname}}} - Uniform (blue), PER (orange) and \emph{\textbf{High \densityname}} - Truncated Geometric (green), and ERE (pink). Solid lines show the mean goal online over all available seeds for each method, and shaded regions indicate $95\%$ bootstrap confidence intervals computed with \texttt{rliable}.}
    \label{fig:Meta-World-multitask-pertask-1}
\end{figure}
\newpage
\begin{figure}[H]
    \centering
    \includegraphics[height=\textheight ,width=\linewidth, keepaspectratio]{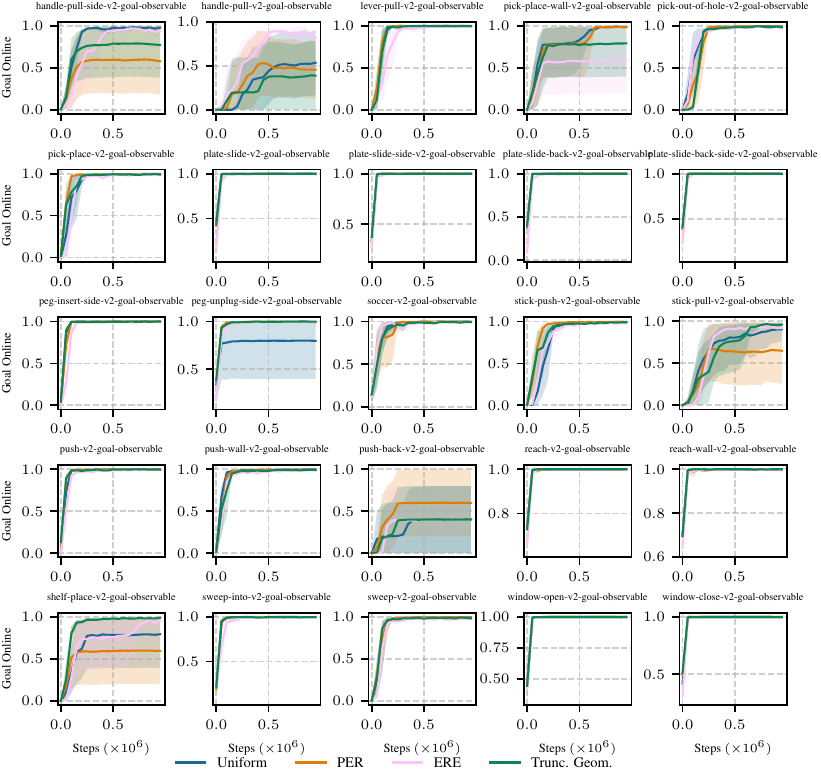}
    \caption{\emph{\textbf{(Low \text{\Volumename}) Mean performance of BRC~\citep{nauman2025bigger} with various replay strategies on Meta-World.} }
    We use \emph{\textbf{Low \text{\densityname}}} - Uniform (blue), PER (orange) and \emph{\textbf{High \densityname}} - Truncated Geometric (green), and ERE (pink). Solid lines show the mean goal online over all available seeds for each method, and shaded regions indicate $95\%$ bootstrap confidence intervals computed with \texttt{rliable}.}
    \label{fig:Meta-World-multitask-pertask-2}
\end{figure}
\newpage
\subsection{FastTD3 Isaac Lab}
\label{app:fasttd3-isaac}

\begin{figure}[H]
    \centering
    \includegraphics{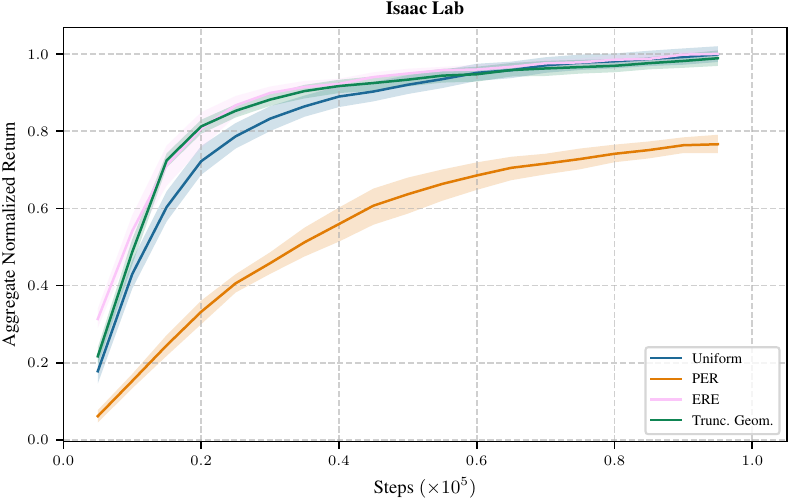}
    \caption{\emph{\textbf{(Low \text{\Volumename}, High \text{\Densityname}) Mean performance of FastTD3~\citep{FastTD3} with various replay strategies on Isaac Lab.}} In this setting due to low buffer size all replay schemes produce high \text{\densityname} replay. Solid lines show mean returns over 5 seeds normalized for each task by dividing returns by mean return from final timestep from uniform sampling strategy. Shaded regions indicate $95\%$ bootstrap confidence intervals computed with \texttt{rliable}~\citep{RLiable}. Truncated Geometric performs comparably to ERE in aggregate, with ERE benefiting from particularly strong performance on one task.}
    \label{fig:isaaclab-fasttd3}
\end{figure}
\begin{figure}[H]
    \centering
    \includegraphics{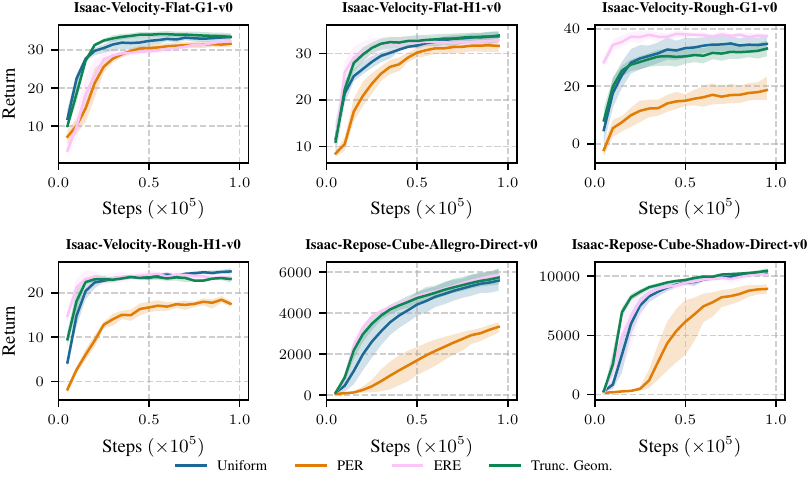}
    \caption{\emph{\textbf{(Low \text{\Volumename}, High \text{\Densityname}) Mean performance of FastTD3 on various replay strategies for each Isaac Lab task.}} Solid lines show mean returns over 5 seeds. Shaded regions indicate $95\%$ bootstrap confidence intervals computed with \texttt{rliable}.}
    \label{fig:isaaclab-fasttd3-pertask}
\end{figure}

\newpage
\subsection{DQN Atari}
\label{app:atari}

\begin{figure}[H]
    \centering
    \includegraphics[width=\linewidth]{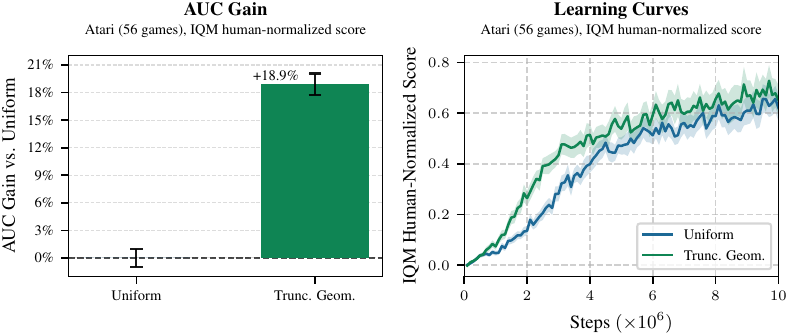}
    \caption{\emph{\textbf{(Low Replay Volume) DQN on Atari (56 games): Truncated Geometric vs.\ Uniform replay.}}
    \emph{Left:} AUC gain over Uniform, where AUC is the time-average of the IQM
    learning curve; Truncated Geometric improves AUC by $18.9\%$.
    \emph{Right:} IQM human-normalized score over $56$ games $\times$ $5$ seeds.
    Following standard practice~\citep{RLiable}, we report the interquartile
    mean (IQM), which is robust to outlier games whose normalized scores dominate the mean.
    Error bars and shaded regions show $95\%$ stratified bootstrap CIs computed with
    \texttt{rliable}.}
    \label{fig:atari}
\end{figure}

\newpage
\begin{figure}[H]
    \centering
    \includegraphics[width=\linewidth]{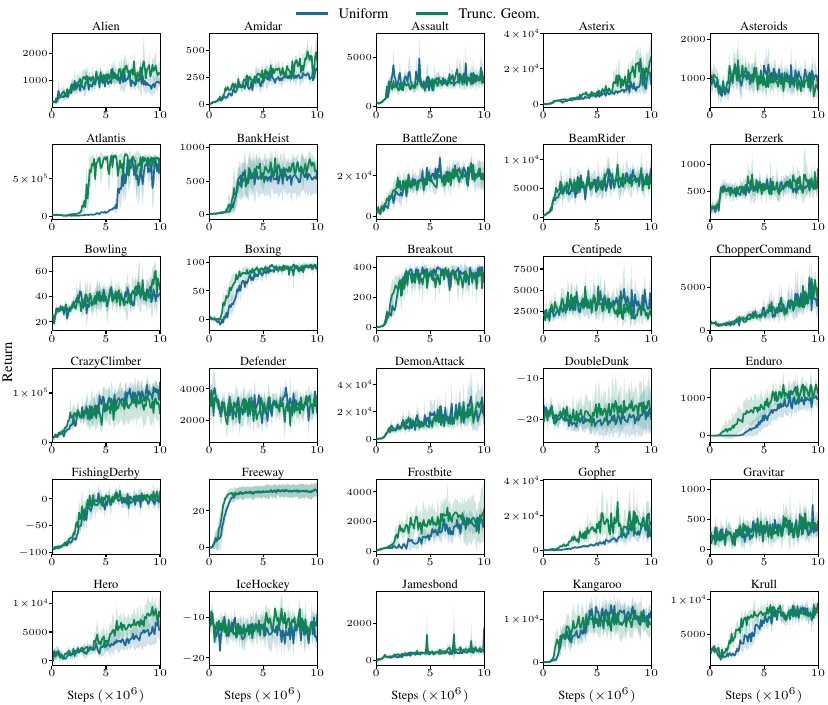}
    \caption{\emph{\textbf{(Low \text{\Volumename}) Per-game performance of DQN~\citep{mnih2015human} with two replay strategies on Atari (Part 1).}}
    We use \emph{\textbf{Low \text{\densityname}}} - Uniform (blue) and \emph{\textbf{High \densityname}} - Truncated Geometric (green). Solid lines show the mean unclipped episodic return over $5$ seeds for each method, and shaded regions indicate $95\%$ bootstrap confidence intervals computed with \texttt{rliable}.}
    \label{fig:atari_per_task1}
\end{figure}

\begin{figure}[H]
    \centering
    \includegraphics{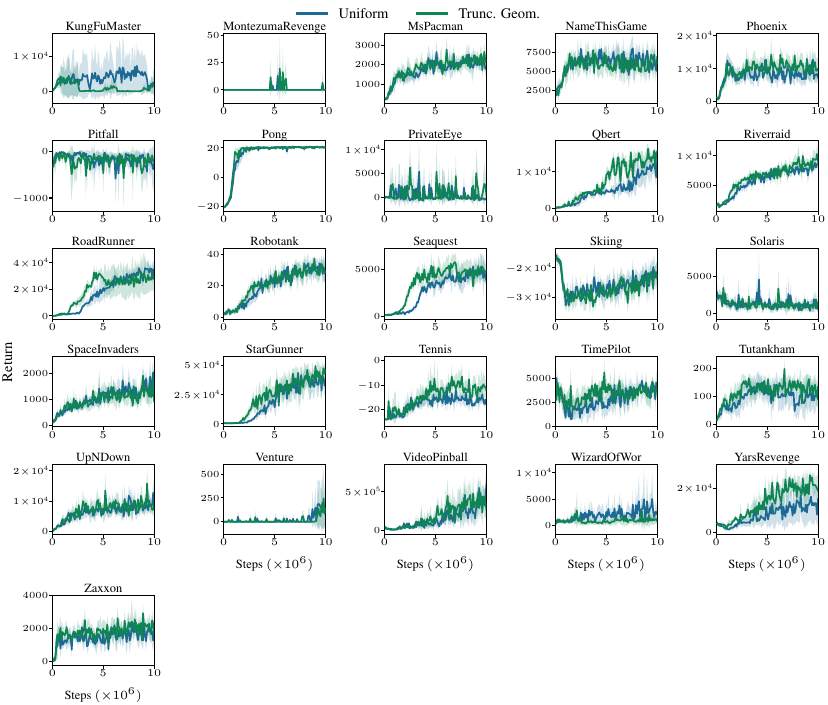}
    \caption{\emph{\textbf{(Low \text{\Volumename}) Per-game performance of DQN~\citep{mnih2015human} with two replay strategies on Atari (Part 2).}}
    We use \emph{\textbf{Low \text{\densityname}}} - Uniform (blue) and \emph{\textbf{High \densityname}} - Truncated Geometric (green). Solid lines show the mean unclipped episodic return over $5$ seeds for each method, and shaded regions indicate $95\%$ bootstrap confidence intervals computed with \texttt{rliable}.}
    \label{fig:atari_per_task2}
\end{figure}

\subsection{Constant replay volume on SimbaV2}
\label{app:constant-replay-simba}
\begin{table}[H]
    \centering
    \begin{tikzpicture}
    \matrix (m) [matrix of nodes, ampersand replacement=\&,
        nodes={font=\tabFont, text height=1.7ex, text depth=0.5ex,
               inner xsep=\tabColPad, inner ysep=\tabRowPad,
               align=right},
        column 1/.style={nodes={text width=0.9cm, align=left}},
        column 2/.style={nodes={text width=1.6cm}},
        column 3/.style={nodes={text width=2.6cm}},
        column 4/.style={nodes={text width=2.6cm}},
        column 5/.style={nodes={text width=1.4cm}},
        column 6/.style={nodes={text width=3.0cm}},
        column sep=0pt, row sep=0pt]
    {
      UTD \& Batch size \& Uniform AUC \& TG AUC \& Gap \& AUC gain \\
      2 \& 32 \& 0.528{\scriptsize\,[0.522,\,0.534]} \& 0.574{\scriptsize\,[0.567,\,0.580]} \& $+0.045$ \& $+8.6\%${\scriptsize\,[+6.1,\,+11.3]} \\
      1 \& 64 \& 0.338{\scriptsize\,[0.333,\,0.344]} \& 0.381{\scriptsize\,[0.374,\,0.387]} \& $+0.042$ \& $+12.4\%${\scriptsize\,[+8.8,\,+16.2]} \\
      $\tfrac{1}{2}$ \& 128 \& 0.367{\scriptsize\,[0.362,\,0.372]} \& 0.402{\scriptsize\,[0.395,\,0.408]} \& $+0.035$ \& $+9.5\%${\scriptsize\,[+6.2,\,+12.9]} \\
      $\tfrac{1}{4}$ \& 256 \& 0.326{\scriptsize\,[0.321,\,0.331]} \& 0.377{\scriptsize\,[0.372,\,0.383]} \& $+0.052$ \& $\mathbf{+15.9}\%${\scriptsize\,[+12.3,\,+19.5]} \\
    };
    \draw[thick] (m-1-1.north west) -- (m-1-6.north east);
    \draw (m-1-1.south west) -- (m-1-6.south east);
    \draw[thick] (m-5-1.south west) -- (m-5-6.south east);
    \end{tikzpicture}
    \caption{\textbf{Matched replay volume.} AUC of SimbaV2 with Uniform and Truncated Geometric (TG, $\alpha = 10$) sampling on HumanoidBench (13 tasks, 5 seeds each), at four UTD/batch-size settings with the same number of replayed samples per environment step ($\text{UTD} \times \text{BS} = 64$). AUC is the normalized return averaged over evaluations up to 1M agent steps. Gap $=$ TG $-$ Uniform; AUC gain is relative to Uniform. Brackets show 95\% stratified bootstrap CIs. Action repeat equals the UTD, so the UTD~2 setting repeats each action twice.}
    \label{tab:matched_replay}
\end{table}

\subsection{Stochastic Action SimbaV2}
\label{app:noisy_simba}
\begin{figure}[H]
    \centering
    \includegraphics{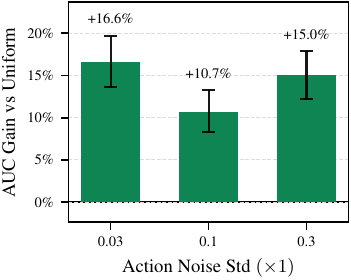}
    \caption{\emph{\textbf{(Low Replay Volume) Noisy SimbaV2 on HumanoidBench (13 tasks).}} Improvement of Truncated Geometric replay over uniform replay at $\text{UTD}=1/4$, $\text{BS}=256$ in stochastic env SimbaV2. Stochasticity is introduced by adding zero-mean Gaussian noise with a given standard deviation to every action before it is executed in the environment (actions are normalized to $[-1, 1]$ and clipped back to that range after the noise is added).
    We report a gain as a ratio of normalized AUCs from normalized returns over $3$ seeds for each method.
    Error bars show $95\%$ stratified bootstrap CIs computed with
    \texttt{rliable.}}
    \label{fig:noise_simba}
\end{figure}

\newpage
\begin{figure}[H]
    \centering
    \includegraphics[width=\linewidth]{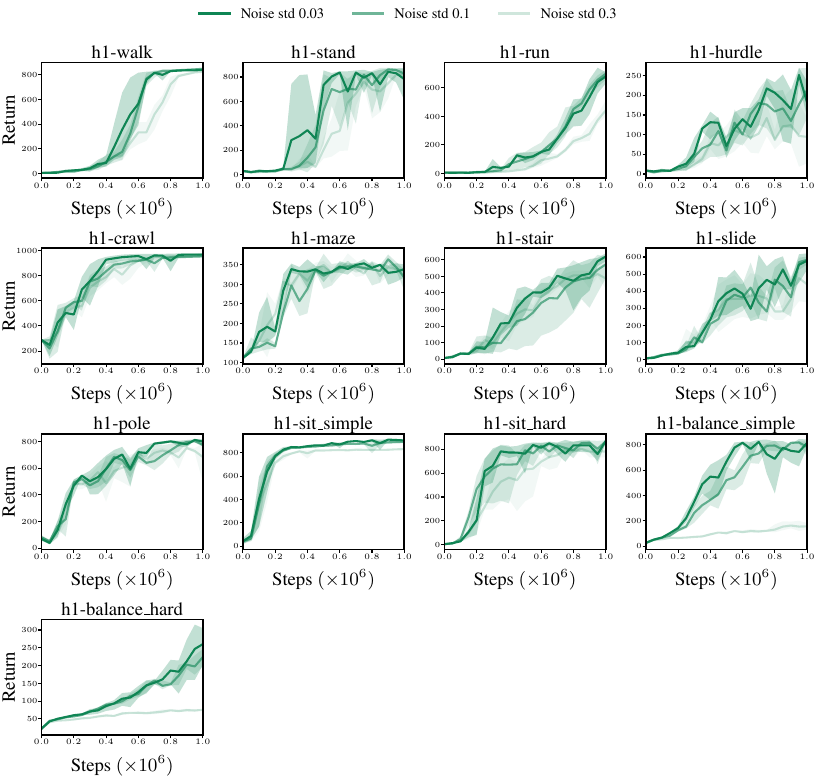}
    \caption{\emph{\textbf{(Low \text{\Volumename}, High \text{\Densityname}) Performance of noisy SimbaV2 with Truncated Geometric replay on HumanoidBench Nohands tasks.}} We report performance for all noise stds from Figure~\ref{fig:noise_simba}. Solid lines show the mean return over all available seeds for each method, and shaded regions indicate $95\%$ bootstrap confidence intervals computed with \texttt{rliable}. All x-axes represent environment steps ($\times 10^6$), and y-axes show the denormalized task returns.}
    \label{fig:hbench-noise-tg-pertask}
\end{figure}

\newpage
\begin{figure}[H]
    \centering
    \includegraphics[width=\linewidth]{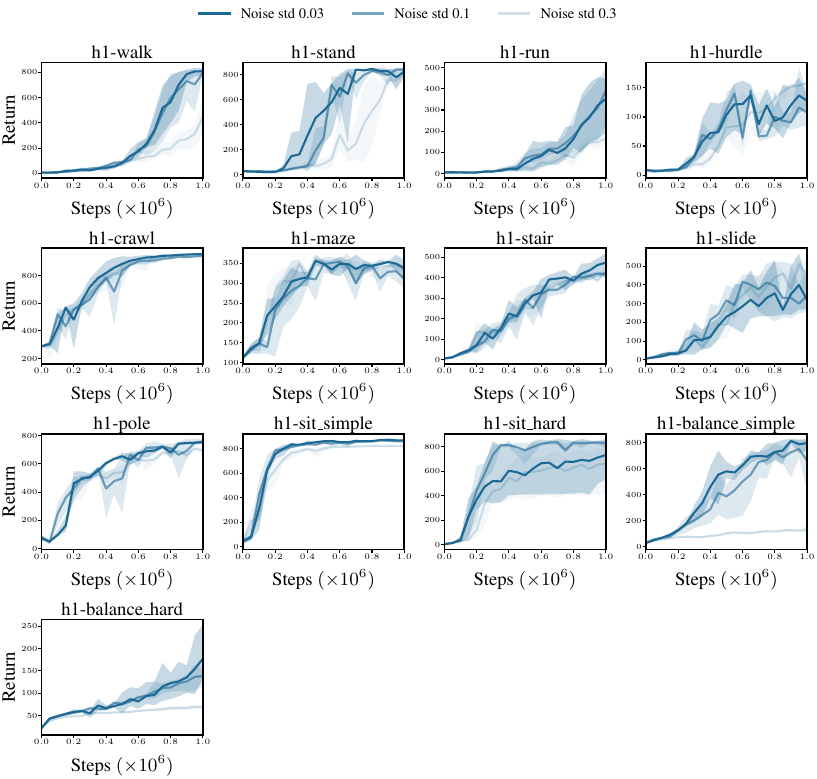}
    \caption{\emph{\textbf{(Low \text{\Volumename}, Low \text{\Densityname}) Performance of noisy SimbaV2 with uniform replay on HumanoidBench Nohands tasks.}} We report performance for all noise stds from Figure~\ref{fig:noise_simba}. Solid lines show the mean return over all available seeds for each method, and shaded regions indicate $95\%$ bootstrap confidence intervals computed with \texttt{rliable}. All x-axes represent environment steps ($\times 10^6$), and y-axes show the denormalized task returns.}
    \label{fig:hbench-noise-uniform-pertask}
\end{figure}

\clearpage
\newpage
\section*{NeurIPS Paper Checklist}

\begin{enumerate}

\item {\bf Claims}
    \item[] Question: Do the main claims made in the abstract and introduction accurately reflect the paper's contributions and scope?
    \item[] Answer: \answerYes{} 
    \item[] Justification: The abstract and introduction clearly state the paper's contributions and scope, as detailed in Section~\ref{sec:introduction} and further supported by the experimental results in Section~\ref{sec:analysis} and Section~\ref{sec:experiments}.
    \item[] Guidelines:
    \begin{itemize}
        \item The answer \answerNA{} means that the abstract and introduction do not include the claims made in the paper.
        \item The abstract and/or introduction should clearly state the claims made, including the contributions made in the paper and important assumptions and limitations. A \answerNo{} or \answerNA{} answer to this question will not be perceived well by the reviewers. 
        \item The claims made should match theoretical and experimental results, and reflect how much the results can be expected to generalize to other settings. 
        \item It is fine to include aspirational goals as motivation as long as it is clear that these goals are not attained by the paper. 
    \end{itemize}

\item {\bf Limitations}
    \item[] Question: Does the paper discuss the limitations of the work performed by the authors?
    \item[] Answer: \answerYes{} 
    \item[] Justification: Limitations are discussed in the appendix~\ref{app:limitations}, acknowledging that experiments focus on continuous control settings and identifying a formal theoretical framework as a promising direction for future work.
    \item[] Guidelines:
    \begin{itemize}
        \item The answer \answerNA{} means that the paper has no limitation while the answer \answerNo{} means that the paper has limitations, but those are not discussed in the paper. 
        \item The authors are encouraged to create a separate ``Limitations'' section in their paper.
        \item The paper should point out any strong assumptions and how robust the results are to violations of these assumptions (e.g., independence assumptions, noiseless settings, model well-specification, asymptotic approximations only holding locally). The authors should reflect on how these assumptions might be violated in practice and what the implications would be.
        \item The authors should reflect on the scope of the claims made, e.g., if the approach was only tested on a few datasets or with a few runs. In general, empirical results often depend on implicit assumptions, which should be articulated.
        \item The authors should reflect on the factors that influence the performance of the approach. For example, a facial recognition algorithm may perform poorly when image resolution is low or images are taken in low lighting. Or a speech-to-text system might not be used reliably to provide closed captions for online lectures because it fails to handle technical jargon.
        \item The authors should discuss the computational efficiency of the proposed algorithms and how they scale with dataset size.
        \item If applicable, the authors should discuss possible limitations of their approach to address problems of privacy and fairness.
        \item While the authors might fear that complete honesty about limitations might be used by reviewers as grounds for rejection, a worse outcome might be that reviewers discover limitations that aren't acknowledged in the paper. The authors should use their best judgment and recognize that individual actions in favor of transparency play an important role in developing norms that preserve the integrity of the community. Reviewers will be specifically instructed to not penalize honesty concerning limitations.
    \end{itemize}

\item {\bf Theory assumptions and proofs}
    \item[] Question: For each theoretical result, does the paper provide the full set of assumptions and a complete (and correct) proof?
    \item[] Answer: \answerNA{} 
    \item[] Justification: The paper is empirical and does not include novel theoretical results; the Truncated Geometric Pareto frontier referenced in Section~\ref{sec:truncated-geometric} is a known result with citations provided.
    \item[] Guidelines:
    \begin{itemize}
        \item The answer \answerNA{} means that the paper does not include theoretical results. 
        \item All the theorems, formulas, and proofs in the paper should be numbered and cross-referenced.
        \item All assumptions should be clearly stated or referenced in the statement of any theorems.
        \item The proofs can either appear in the main paper or the supplemental material, but if they appear in the supplemental material, the authors are encouraged to provide a short proof sketch to provide intuition. 
        \item Inversely, any informal proof provided in the core of the paper should be complemented by formal proofs provided in appendix or supplemental material.
        \item Theorems and Lemmas that the proof relies upon should be properly referenced. 
    \end{itemize}

    \item {\bf Experimental result reproducibility}
    \item[] Question: Does the paper fully disclose all the information needed to reproduce the main experimental results of the paper to the extent that it affects the main claims and/or conclusions of the paper (regardless of whether the code and data are provided or not)?
    \item[] Answer: \answerYes{} 
    \item[] Justification: All information needed to reproduce the experiments is provided, including the algorithms used, benchmark environments (Section~\ref{sec:experiments}), pseudocode of the proposed method (Section~\ref{sec:truncated-geometric}), and hyperparameters detailed in the appendix~\ref{app:hyperparams}.
    \item[] Guidelines:
    \begin{itemize}
        \item The answer \answerNA{} means that the paper does not include experiments.
        \item If the paper includes experiments, a \answerNo{} answer to this question will not be perceived well by the reviewers: Making the paper reproducible is important, regardless of whether the code and data are provided or not.
        \item If the contribution is a dataset and\slash or model, the authors should describe the steps taken to make their results reproducible or verifiable. 
        \item Depending on the contribution, reproducibility can be accomplished in various ways. For example, if the contribution is a novel architecture, describing the architecture fully might suffice, or if the contribution is a specific model and empirical evaluation, it may be necessary to either make it possible for others to replicate the model with the same dataset, or provide access to the model. In general. releasing code and data is often one good way to accomplish this, but reproducibility can also be provided via detailed instructions for how to replicate the results, access to a hosted model (e.g., in the case of a large language model), releasing of a model checkpoint, or other means that are appropriate to the research performed.
        \item While NeurIPS does not require releasing code, the conference does require all submissions to provide some reasonable avenue for reproducibility, which may depend on the nature of the contribution. For example
        \begin{enumerate}
            \item If the contribution is primarily a new algorithm, the paper should make it clear how to reproduce that algorithm.
            \item If the contribution is primarily a new model architecture, the paper should describe the architecture clearly and fully.
            \item If the contribution is a new model (e.g., a large language model), then there should either be a way to access this model for reproducing the results or a way to reproduce the model (e.g., with an open-source dataset or instructions for how to construct the dataset).
            \item We recognize that reproducibility may be tricky in some cases, in which case authors are welcome to describe the particular way they provide for reproducibility. In the case of closed-source models, it may be that access to the model is limited in some way (e.g., to registered users), but it should be possible for other researchers to have some path to reproducing or verifying the results.
        \end{enumerate}
    \end{itemize}

\item {\bf Open access to data and code}
    \item[] Question: Does the paper provide open access to the data and code, with sufficient instructions to faithfully reproduce the main experimental results, as described in supplemental material?
    \item[] Answer: \answerYes{} 
    \item[] Justification: Code is provided in the supplementary material and experiments are conducted on publicly available open-source reinforcement learning environments.
    \item[] Guidelines:
    \begin{itemize}
        \item The answer \answerNA{} means that paper does not include experiments requiring code.
        \item Please see the NeurIPS code and data submission guidelines (\url{https://neurips.cc/public/guides/CodeSubmissionPolicy}) for more details.
        \item While we encourage the release of code and data, we understand that this might not be possible, so \answerNo{} is an acceptable answer. Papers cannot be rejected simply for not including code, unless this is central to the contribution (e.g., for a new open-source benchmark).
        \item The instructions should contain the exact command and environment needed to run to reproduce the results. See the NeurIPS code and data submission guidelines (\url{https://neurips.cc/public/guides/CodeSubmissionPolicy}) for more details.
        \item The authors should provide instructions on data access and preparation, including how to access the raw data, preprocessed data, intermediate data, and generated data, etc.
        \item The authors should provide scripts to reproduce all experimental results for the new proposed method and baselines. If only a subset of experiments are reproducible, they should state which ones are omitted from the script and why.
        \item At submission time, to preserve anonymity, the authors should release anonymized versions (if applicable).
        \item Providing as much information as possible in supplemental material (appended to the paper) is recommended, but including URLs to data and code is permitted.
    \end{itemize}

\item {\bf Experimental setting/details}
    \item[] Question: Does the paper specify all the training and test details (e.g., data splits, hyperparameters, how they were chosen, type of optimizer) necessary to understand the results?
    \item[] Answer: \answerYes{} 
    \item[] Justification: Full training details, hyperparameters, evaluation protocols, and metrics are provided in the appendix~\ref{app:training_details}.
    \item[] Guidelines:
    \begin{itemize}
        \item The answer \answerNA{} means that the paper does not include experiments.
        \item The experimental setting should be presented in the core of the paper to a level of detail that is necessary to appreciate the results and make sense of them.
        \item The full details can be provided either with the code, in appendix, or as supplemental material.
    \end{itemize}

\item {\bf Experiment statistical significance}
    \item[] Question: Does the paper report error bars suitably and correctly defined or other appropriate information about the statistical significance of the experiments?
    \item[] Answer: \answerYes{} 
    \item[] Justification: All results report 95\% confidence intervals computed via stratified bootstrap resampling over runs following the RLiable~\citep{RLiable} protocol, with the aggregation procedure detailed in the appendix~\ref{app:auc}.
    \item[] Guidelines:
    \begin{itemize}
        \item The answer \answerNA{} means that the paper does not include experiments.
        \item The authors should answer \answerYes{} if the results are accompanied by error bars, confidence intervals, or statistical significance tests, at least for the experiments that support the main claims of the paper.
        \item The factors of variability that the error bars are capturing should be clearly stated (for example, train/test split, initialization, random drawing of some parameter, or overall run with given experimental conditions).
        \item The method for calculating the error bars should be explained (closed form formula, call to a library function, bootstrap, etc.)
        \item The assumptions made should be given (e.g., Normally distributed errors).
        \item It should be clear whether the error bar is the standard deviation or the standard error of the mean.
        \item It is OK to report 1-sigma error bars, but one should state it. The authors should preferably report a 2-sigma error bar than state that they have a 96\% CI, if the hypothesis of Normality of errors is not verified.
        \item For asymmetric distributions, the authors should be careful not to show in tables or figures symmetric error bars that would yield results that are out of range (e.g., negative error rates).
        \item If error bars are reported in tables or plots, the authors should explain in the text how they were calculated and reference the corresponding figures or tables in the text.
    \end{itemize}

\item {\bf Experiments compute resources}
    \item[] Question: For each experiment, does the paper provide sufficient information on the computer resources (type of compute workers, memory, time of execution) needed to reproduce the experiments?
    \item[] Answer: \answerYes{} 
    \item[] Justification:  Computational resources are detailed in Appendix~\ref{app:resources}, including GPU type and estimated runtimes for single-task, large scale parallel and multi-task experiments.
    \item[] Guidelines:
    \begin{itemize}
        \item The answer \answerNA{} means that the paper does not include experiments.
        \item The paper should indicate the type of compute workers CPU or GPU, internal cluster, or cloud provider, including relevant memory and storage.
        \item The paper should provide the amount of compute required for each of the individual experimental runs as well as estimate the total compute. 
        \item The paper should disclose whether the full research project required more compute than the experiments reported in the paper (e.g., preliminary or failed experiments that didn't make it into the paper). 
    \end{itemize}
    
\item {\bf Code of ethics}
    \item[] Question: Does the research conducted in the paper conform, in every respect, with the NeurIPS Code of Ethics \url{https://neurips.cc/public/EthicsGuidelines}?
    \item[] Answer: \answerYes{} 
    \item[] Justification: The research fully conforms with the NeurIPS Code of Ethics. It involves no human subjects, sensitive data, or foreseeable harmful applications.
    \item[] Guidelines:
    \begin{itemize}
        \item The answer \answerNA{} means that the authors have not reviewed the NeurIPS Code of Ethics.
        \item If the authors answer \answerNo, they should explain the special circumstances that require a deviation from the Code of Ethics.
        \item The authors should make sure to preserve anonymity (e.g., if there is a special consideration due to laws or regulations in their jurisdiction).
    \end{itemize}

\item {\bf Broader impacts}
    \item[] Question: Does the paper discuss both potential positive societal impacts and negative societal impacts of the work performed?
    \item[] Answer: \answerNA{} 
    \item[] Justification: The paper presents foundational research on replay sampling strategies for reinforcement learning, with no direct path to negative societal impacts or specific real-world deployment concerns.
    \item[] Guidelines:
    \begin{itemize}
        \item The answer \answerNA{} means that there is no societal impact of the work performed.
        \item If the authors answer \answerNA{} or \answerNo, they should explain why their work has no societal impact or why the paper does not address societal impact.
        \item Examples of negative societal impacts include potential malicious or unintended uses (e.g., disinformation, generating fake profiles, surveillance), fairness considerations (e.g., deployment of technologies that could make decisions that unfairly impact specific groups), privacy considerations, and security considerations.
        \item The conference expects that many papers will be foundational research and not tied to particular applications, let alone deployments. However, if there is a direct path to any negative applications, the authors should point it out. For example, it is legitimate to point out that an improvement in the quality of generative models could be used to generate Deepfakes for disinformation. On the other hand, it is not needed to point out that a generic algorithm for optimizing neural networks could enable people to train models that generate Deepfakes faster.
        \item The authors should consider possible harms that could arise when the technology is being used as intended and functioning correctly, harms that could arise when the technology is being used as intended but gives incorrect results, and harms following from (intentional or unintentional) misuse of the technology.
        \item If there are negative societal impacts, the authors could also discuss possible mitigation strategies (e.g., gated release of models, providing defenses in addition to attacks, mechanisms for monitoring misuse, mechanisms to monitor how a system learns from feedback over time, improving the efficiency and accessibility of ML).
    \end{itemize}
    
\item {\bf Safeguards}
    \item[] Question: Does the paper describe safeguards that have been put in place for responsible release of data or models that have a high risk for misuse (e.g., pre-trained language models, image generators, or scraped datasets)?
    \item[] Answer: \answerNA{} 
    \item[] Justification: The paper releases only reinforcement learning code and poses no risk of misuse; no safeguards are required.
    \item[] Guidelines:
    \begin{itemize}
        \item The answer \answerNA{} means that the paper poses no such risks.
        \item Released models that have a high risk for misuse or dual-use should be released with necessary safeguards to allow for controlled use of the model, for example by requiring that users adhere to usage guidelines or restrictions to access the model or implementing safety filters. 
        \item Datasets that have been scraped from the Internet could pose safety risks. The authors should describe how they avoided releasing unsafe images.
        \item We recognize that providing effective safeguards is challenging, and many papers do not require this, but we encourage authors to take this into account and make a best faith effort.
    \end{itemize}

\item {\bf Licenses for existing assets}
    \item[] Question: Are the creators or original owners of assets (e.g., code, data, models), used in the paper, properly credited and are the license and terms of use explicitly mentioned and properly respected?
    \item[] Answer: \answerYes{} 
    \item[] Justification: All benchmarks used are open-source and properly cited in Section~\ref{sec:experiments}; all models are trained from scratch with no additional assets requiring licensing.
    \item[] Guidelines:
    \begin{itemize}
        \item The answer \answerNA{} means that the paper does not use existing assets.
        \item The authors should cite the original paper that produced the code package or dataset.
        \item The authors should state which version of the asset is used and, if possible, include a URL.
        \item The name of the license (e.g., CC-BY 4.0) should be included for each asset.
        \item For scraped data from a particular source (e.g., website), the copyright and terms of service of that source should be provided.
        \item If assets are released, the license, copyright information, and terms of use in the package should be provided. For popular datasets, \url{paperswithcode.com/datasets} has curated licenses for some datasets. Their licensing guide can help determine the license of a dataset.
        \item For existing datasets that are re-packaged, both the original license and the license of the derived asset (if it has changed) should be provided.
        \item If this information is not available online, the authors are encouraged to reach out to the asset's creators.
    \end{itemize}

\item {\bf New assets}
    \item[] Question: Are new assets introduced in the paper well documented and is the documentation provided alongside the assets?
    \item[] Answer: \answerYes{} 
    \item[] Justification: The code released alongside the paper is documented and provided as an anonymized zip file in the supplementary material.
    \item[] Guidelines:
    \begin{itemize}
        \item The answer \answerNA{} means that the paper does not release new assets.
        \item Researchers should communicate the details of the dataset\slash code\slash model as part of their submissions via structured templates. This includes details about training, license, limitations, etc. 
        \item The paper should discuss whether and how consent was obtained from people whose asset is used.
        \item At submission time, remember to anonymize your assets (if applicable). You can either create an anonymized URL or include an anonymized zip file.
    \end{itemize}

\item {\bf Crowdsourcing and research with human subjects}
    \item[] Question: For crowdsourcing experiments and research with human subjects, does the paper include the full text of instructions given to participants and screenshots, if applicable, as well as details about compensation (if any)? 
    \item[] Answer: \answerNA{} 
    \item[] Justification:  The paper does not involve crowdsourcing or research with human subjects.
    \item[] Guidelines:
    \begin{itemize}
        \item The answer \answerNA{} means that the paper does not involve crowdsourcing nor research with human subjects.
        \item Including this information in the supplemental material is fine, but if the main contribution of the paper involves human subjects, then as much detail as possible should be included in the main paper. 
        \item According to the NeurIPS Code of Ethics, workers involved in data collection, curation, or other labor should be paid at least the minimum wage in the country of the data collector. 
    \end{itemize}

\item {\bf Institutional review board (IRB) approvals or equivalent for research with human subjects}
    \item[] Question: Does the paper describe potential risks incurred by study participants, whether such risks were disclosed to the subjects, and whether Institutional Review Board (IRB) approvals (or an equivalent approval/review based on the requirements of your country or institution) were obtained?
    \item[] Answer: \answerNA{} 
    \item[] Justification: The paper does not involve human subjects and therefore requires no IRB approval.
    \item[] Guidelines:
    \begin{itemize}
        \item The answer \answerNA{} means that the paper does not involve crowdsourcing nor research with human subjects.
        \item Depending on the country in which research is conducted, IRB approval (or equivalent) may be required for any human subjects research. If you obtained IRB approval, you should clearly state this in the paper. 
        \item We recognize that the procedures for this may vary significantly between institutions and locations, and we expect authors to adhere to the NeurIPS Code of Ethics and the guidelines for their institution. 
        \item For initial submissions, do not include any information that would break anonymity (if applicable), such as the institution conducting the review.
    \end{itemize}

\item {\bf Declaration of LLM usage}
    \item[] Question: Does the paper describe the usage of LLMs if it is an important, original, or non-standard component of the core methods in this research? Note that if the LLM is used only for writing, editing, or formatting purposes and does \emph{not} impact the core methodology, scientific rigor, or originality of the research, declaration is not required.
    \item[] Answer: \answerNA{} 
    \item[] Justification: LLMs are not used as part of the core methodology; any use was limited to writing and editing purposes only.
    \item[] Guidelines:
    \begin{itemize}
        \item The answer \answerNA{} means that the core method development in this research does not involve LLMs as any important, original, or non-standard components.
        \item Please refer to our LLM policy in the NeurIPS handbook for what should or should not be described.
    \end{itemize}

\end{enumerate}

\end{document}